\documentclass[10pt,journal,compsoc]{IEEEtran}
\ifCLASSOPTIONcompsoc
  \usepackage[nocompress]{cite}
\else
  \usepackage{cite}
\fi
\ifCLASSINFOpdf
\else
\fi
\usepackage{lineno,hyperref}
\modulolinenumbers[5]
\usepackage{graphicx} 
\usepackage{subcaption}
\usepackage{float}
\usepackage{tabularx}
\usepackage{amsmath}
\usepackage{amsfonts}
\usepackage{amsthm}
\usepackage{amssymb}
\usepackage{dsfont}
\usepackage{booktabs}
\usepackage{algorithm}
\usepackage{algorithmic}
\usepackage{booktabs}
\usepackage[makeroom]{cancel}
\usepackage{color}

\newcommand\given [1][]{\:#1\vert\:}

\begin{document}
%
\title{A review of domain adaptation\\ without target labels}
%
%
%
%

\author{Wouter~M.~Kouw, 
        ~Marco~Loog
\IEEEcompsocitemizethanks{\IEEEcompsocthanksitem W.M. Kouw is with the Datalogisk Institut, University of Copenhagen, at Universitetsparken 1, DK-2100 Copenhagen, Denmark.\protect\\
E-mail: wmkouw@gmail.com
\IEEEcompsocthanksitem M. Loog is with the Pattern Recognition Laboratory, Delft University of Technology, in Delft, the Netherlands and the Datalogisk Institut, University of Copenhagen, in Copenhagen, Denmark.}
\thanks{}}

%
%

\markboth{}%
{Kouw \& Loog: A review of domain adaptation without target labels.}
\IEEEtitleabstractindextext{%
\begin{abstract}
Domain adaptation has become a prominent problem setting in machine learning and related fields. This review asks the question:  \emph{how} can a classifier learn from a source domain and generalize to a target domain? We present a categorization of approaches, divided into, what we refer to as, \emph{sample-based}, \emph{feature-based} and \emph{inference-based} methods. Sample-based methods focus on weighting individual observations during training based on their importance to the target domain. Feature-based methods revolve around on mapping, projecting and representing features such that a source classifier performs well on the target domain and inference-based methods incorporate adaptation into the parameter estimation procedure, for instance through constraints on the optimization procedure. Additionally, we review a number of conditions that allow for formulating bounds on the cross-domain generalization error. Our categorization highlights recurring ideas and raises questions important to further research.
\end{abstract}

\begin{IEEEkeywords}
Machine Learning, Pattern Recognition, Domain Adaptation, Transfer Learning, Covariate Shift, Sample Selection Bias.
\end{IEEEkeywords}}

\maketitle

\IEEEdisplaynontitleabstractindextext

\IEEEpeerreviewmaketitle

\IEEEraisesectionheading{\section{Introduction}\label{sec:introduction}}
\IEEEPARstart{G}{eneralization} is the process of observing a finite number of samples and making statements about all possible samples. In the case of machine learning and pattern recognition, samples are used to train classifiers to make predictions for future samples. However, if the observed labeled samples are \emph{not} an accurate reflection of the underlying distribution on which the learner should operate, the system will \emph{not} generalize well to new samples. In practice, collected data is hardly ever a completely unbiased representation of the operating setting.

Data is \emph{biased} if certain outcomes are systematically more frequently observed  than they would be for a uniformly-at-random sampling procedure. For example, data sampled from a single hospital can be biased with respect to the global population due to differences in living conditions of the local patient population. Statisticians have long studied sampling biases under the term \emph{sample selection bias} \cite{heckman1977sample,heckman1990varieties}. Corrections are based on estimating - or in cases where there is control of the experimental design, \emph{knowing} - the probability that an instance will be selected for observation \cite{cochran1973controlling,little2014statistical,imbens2015causal}. But many modern data collection procedures, such as internet crawlers, are less structured than sampling from patients in a hospital, for instance. It is therefore difficult, if not impossible, to estimate the probability that a sample is selected for observation and by extension, how the biased sample differs from the general population. On the other hand, it might not be necessary to generalize to the whole population. It could be more important to generalize to a specific \emph{target} subpopulation. For example, can data collected in European hospitals be used to train an intelligent prognosis system for hospitals in Africa?

In order to target specific distributions over sample space, henceforth referred to as \emph{domains}, we need at least \emph{some} information. Unlabeled data from a target domain can usually be collected, but labels are more difficult to obtain. Nevertheless, unlabeled data gives an indication in what way a source domain and a target domain differ from each other. 
This information can be exploited to make a classifier \emph{adapt}, i.e. change its decisions such that it generalizes better towards the target domain.
 
The important question to ask is: \emph{how} can a classifier learn from a source domain and generalize to a target domain? 
We present a categorization of methods into three parts, each containing a subcategorization on a finer level. First, there are \emph{sample-based} methods, which are based on correcting for biases in the data sampling procedure through individual samples. Methods in this category focus on data importance-weighting \cite{zadrozny2004learning,cortes2014domain} or class importance-weighting \cite{lipton2018detecting}. Secondly, there are \emph{feature-based} methods, which focus on reshaping feature space such that a classifier trained on transformed source data will generalize to target data. A further distinction can be made into finding subspace mappings \cite{gong2012geodesic,fernando2013unsupervised}, optimal transportation techniques \cite{courty2016optimal}, learning domain-invariant representations \cite{pan2011domain} or constructing corresponding features \cite{blitzer2006domain}. Thirdly, we consider what we call \emph{inference-based} approaches. These methods focus on incorporating the adaptation into the parameter estimation procedure. It is a diverse category, containing algorithmic robustness \cite{mansour2014robust}, minimax estimators \cite{wen2014robust}, self-learning \cite{bruzzone2010domain}, empirical Bayes \cite{raina2006constructing} and PAC-Bayes \cite{germain2007pac}. Clearly, the above classification is not necessarily mutually exclusive, but we believe it offers a comprehensible overview.

Our categorization reveals a small number of conditions that permit performance guarantees for domain adaptive classifiers. In practice, one has to assume that a condition holds, which means that for any adaptive classifier a problem setting exists for which it fails. We discuss the importance of hypothesis tests and causal information to domain-adaptive classifier selection. 

\subsection{Scope}
Our scope is limited to the single-source $/$ single-target adaptation without labeled data from the target domain. This setting is the minimal form to study cross-domain generalization. Incorporating multiple source domains raises additional questions such as: 
Should each source domain be weighted based on its similarity to the target domain? Or should domains be selected? Should there be some temporal or spatial ordering? Questions concerning multi-source adaptation are not considered here, but the interested reader may refer to \cite{mansour2009domaina,mansour2009multiple,sun2015survey}. Similarly, incorporating target labels would raise additional questions relating to semi-supervised learning, active learning and multi-task learning \cite{ando2005framework,donahue2013semi}. For example, how can the unlabeled target data improve the classifier's estimation? Are some labeled target samples more informative than others? Are there features that are useful to classification in both domains? These questions will remain outside our scope as well. Topics that are well covered by other reviews will not be discussed in great detail here either. First, there are two articles covering visual domain adaptation \cite{patel2015visual,csurka2017domain}, with a third one specializing in deep learning \cite{wang2018deep}. Secondly, there is an empirical comparison of domain adaptation methods for genomic sequence analysis \cite{schweikert2009empirical} and thirdly, a survey paper on, amongst others, transfer learning in biomedical imaging \cite{cheplygina2019not}. 

Other reviews are available: there is a book on data set shift in machine learning \cite{quionero2009dataset}, an excellent paper on the types and causes of data set shift \cite{moreno2012unifying}, a technical report on domain adaptation with unlabeled samples \cite{margolis2011literature} and two papers focusing on variants of transfer learning \cite{pan2010survey,arnold2007comparative}. Our work complements these with more recent studies.

\subsection{Outline}
Before the end of this section, we go through some motivating examples. These demonstrate that this problem is relevant to a wide variety of scientific and engineering fields. In section \ref{sec:problem} we turn to more precise definitions of domains and adaptation. Furthermore, we discuss briefly a number of assumptions that permit performance guarantees on the target domain. Additionally, an example setting is presented that will serve to provide some intuition of how particular types of methods work. Our categorization starts with sample-based methods in Section \ref{sec:sample}. Following that are feature-based methods in Section \ref{sec:feature} and inference-based methods in Section \ref{sec:inference}. The discussion of methods includes equations whenever they facilitate comparisons, such as between divergences or estimators. Lastly, we discuss common themes and open questions in Section \ref{sec:discussion}, and summarize our findings in Section \ref{sec:conclusion}. 

\subsection{Motivating examples}

Sample selection bias has been studied in statistics and econometrics for quite some time \cite{heckman1990varieties,tucker2010selection}. For example, in the 70s, it was of interest to find predictors for wage rates of women. These were estimated by measuring characteristics of working women and their salaries \cite{gronau1974wage,vella1998estimating}. However, working women differed from non-working women in terms such as age, number of children, and education \cite{gronau1974wage}. The predictor did not generalize to the total population, which raised awareness of sample selection bias as an issue.

In clinical studies, randomized controlled trials are used to study the effects of treatments \cite{hernan2006estimating}. 
The average treatment effect is estimated by comparing the difference between an experimental and a control group \cite{imbens2015causal}. The treatment effect is expected to hold for patients outside of the study, but that is not necessarily the case if certain patients were systematically excluded from the study. For example, the factor that makes certain patients non-compliant -- a reason for exclusion -- can also have an effect on the treatment \cite{hernan2006estimating}.

In medical imaging, radiologists manually annotate tissues, abnormalities, and pathologies to obtain training data for computer-aided-diagnosis systems. But due to the mechanical configuration, calibration, vendor or acquisition protocol of MRI, CT or PET scanners, there are large variations between data sets from different medical centers \cite{stonnington2008interpreting,cheplygina2017transfer}. Consequently, diagnosis systems often fail to perform well across centers \cite{van2015transfer,kamnitsas2017unsupervised,kouw2019cross}. 

Computer vision deals with such rich, high-dimensional data that even large image data sets are essentially biased samplings of visual objects \cite{torralba2011unbiased}. As a result, cross-dataset generalization is low for systems that do not employ some form of adaptation \cite{torralba2011unbiased,patel2015visual,csurka2017domain}. Examples of adaptation studies include recognizing objects in photos based on commercial images \cite{saenko2010adapting}, event recognition in consumer videos through training on web data \cite{duan2011visual}, recognizing activities across viewpoints \cite{farhadi2008learning}, and recognizing movements across sensors \cite{van2010transferring} and persons \cite{hachiya2012importance,gedik2017personalised}.

In robotics, simulations can be employed as an additional source of data \cite{tai2017virtual,marco2017virtual}. Physics simulators have been extensively developed for fields such as computer graphics or video gaming and one could potentially generate a vast amount of data. Through adaptation from the simulated data to the real data, domain adaptive methods can be helpful in improving lane and pedestrian detection for self-driving cars \cite{hoffman2016fcns,wulfmeier2017addressing} or improving hand grasping for stationary robots \cite{bousmalis2018using}.

Speech differs strongly across speakers, but is instantly recognizable for humans. Learning algorithms expect new vocal data to be similar to training data, and struggle to generalize across speakers \cite{deng2014autoencoder}. With the adoption of commercial speech recognition systems in homes, there is an increasing need to adapt to specific speakers \cite{wang2018unsupervised}.

In natural language processing, authors are known to use different styles of expression on different publication platforms. For instance, biomedical science articles contain words like 'oncogenic' and 'mutated', which appear far less often in financial news articles \cite{blitzer2006domain}. Similarly, online movie reviews are linguistically different from tweets \cite{peddinti2011domain} and product reviews differ per category \cite{blitzer2007biographies}. Natural language processing tasks such as sentiment classification or named-entity recognition become much more challenging under changes in word frequencies.

In bioinformatics, adaptive approaches have been successful in sequence classification \cite{widmer2010novel,mei2011gene} and biological network reconstruction \cite{kato2012transfer}. For some problem settings, the goal is to generalize from one model organism to another \cite{xu2011survey}, such as from nematodes to fruit flies \cite{schweikert2009empirical}.

Radiotelescopes measure spectral signals arriving to earth. Astronomers use these signals to, for example, detect quasars or to determine the amount of photometric redshift in galaxies \cite{gieseke2010detecting,kremer2015nearest}. These signals are costly to label, and astronomers therefore choose the ones they consider the most promising. But this selection constitutes a biased sampling procedure and various domain adaptation techniques have been employed to tackle it \cite{vilalta2013machine,izbicki2017photo}.


Fairness-aware machine learning focuses on ensuring that algorithms and automated decision-making systems do not discriminate based on gender, race or other protected attributes \cite{hardt2016equality,friedler2019comparative}. For instance, if a data set contains many examples of men with higher salaries, then a recommender system could learn to suggest mostly men as candidates for positions with high salaries. To tackle this kind of unwanted behavior in algorithms, fairness has to be built into the learning process \cite{quadrianto2017recycling}. Aspects of fairness, such as equality of opportunity, can be formulated as constraints on the learning process \cite{hardt2016equality}. Some of these constraints lead to a need for distribution matching techniques, which have been extensively developed for domain adaptation \cite{madras2018learning}.

\section{Domain adaptation} \label{sec:problem}
We go through some definitions relevant to domain adaptation, followed by the introduction of a running example, a remark on domain discrepancy metrics and a brief review of assumptions that permit generalization error bounds for adaptive classifiers. The reader is assumed to be familiar with supervised learning and risk minimization. For extensive overviews of these topics, see \cite{mohri2012foundations,loog2018supervised}.

\subsection{Definitions \& notation}
Consider an input or \emph{feature} space $\mathcal{ X}$, a subset of $\mathbb{R}^{D}$, and an output or \emph{label} space $\mathcal{ Y}$, either binary $\{-1,+1\}$ or multi-class $\{1, \dots K\}$ with $K$ as the number of classes. We define different \emph{domains}, in this context, as different probability distributions $p(x,y)$ over the same feature-label space pair $\mathcal{X} \times \mathcal{Y}$. The domain of interest is called the \emph{target} domain $p_{\cal T}(x,y)$ and the domain with labeled data is called the \emph{source} domain $p_{\cal S}(x,y)$. \emph{Domain adaptation} refers to predicting the labels of samples drawn from a target domain, given labeled samples drawn from a source domain and unlabeled samples drawn from the target domain itself. We consider domain adaptation a special case of \emph{transfer learning}, where differences between feature spaces and label spaces are allowed. For example, transferring from the "speech domain" to the "text domain". 

We assume that a data set of size $n$ from the source domain is given. Source samples are marked $x_i$ and source labels are marked $y_i$,  for $i=1, \dots, n$. Likewise, a data set of size $m$ drawn from the target domain, where only the target samples $z_j$ for $j = 1, \dots, m$ are given and the labels $u_j$ are not known. A problem setting with at least one observed target label is usually called \emph{semi-supervised domain adaptation}. There are two potential goals: firstly, to predict the labels of the \emph{given} target samples. The second goal is to predict the labels of \emph{new} samples from the target domain. The first one is called the \emph{transductive} setting, and the second goal is called the \emph{inductive} setting.

Functions and distributions related to the source domain are marked with the subscript $\mathcal{S}$, e.g. the source posterior distribution $p_{\cal S}(y \given x)$. Similarly, functions related to the target domain are marked with $\mathcal{T}$. Classification functions map samples to a real number, $h : {\cal X} \rightarrow \mathbb{R}$, where $h$ is an element of an hypothesis space ${\cal H}$. A loss function $\ell$ compares a classifier prediction with the true label $\ell : \mathbb{R} \times {\cal Y} \rightarrow \mathbb{R}$. A risk function $R$ is the expected loss of a particular classifier, with respect to a distribution $R(h) = \mathbb{E}[ \ell(h(x), y)]$. The error function $e$ is a special case of a risk function, corresponding to the expected $0$/$1$-loss. $\mathrm{D}$ is reserved for discrepancy measures, $W$ for weights, $M$ for transformation matrices, $\phi$ for basis functions and $\kappa$ for kernel functions.

\subsection{Example setting} \label{sec:example}
\begin{figure}[thb]
\includegraphics[height=120px]{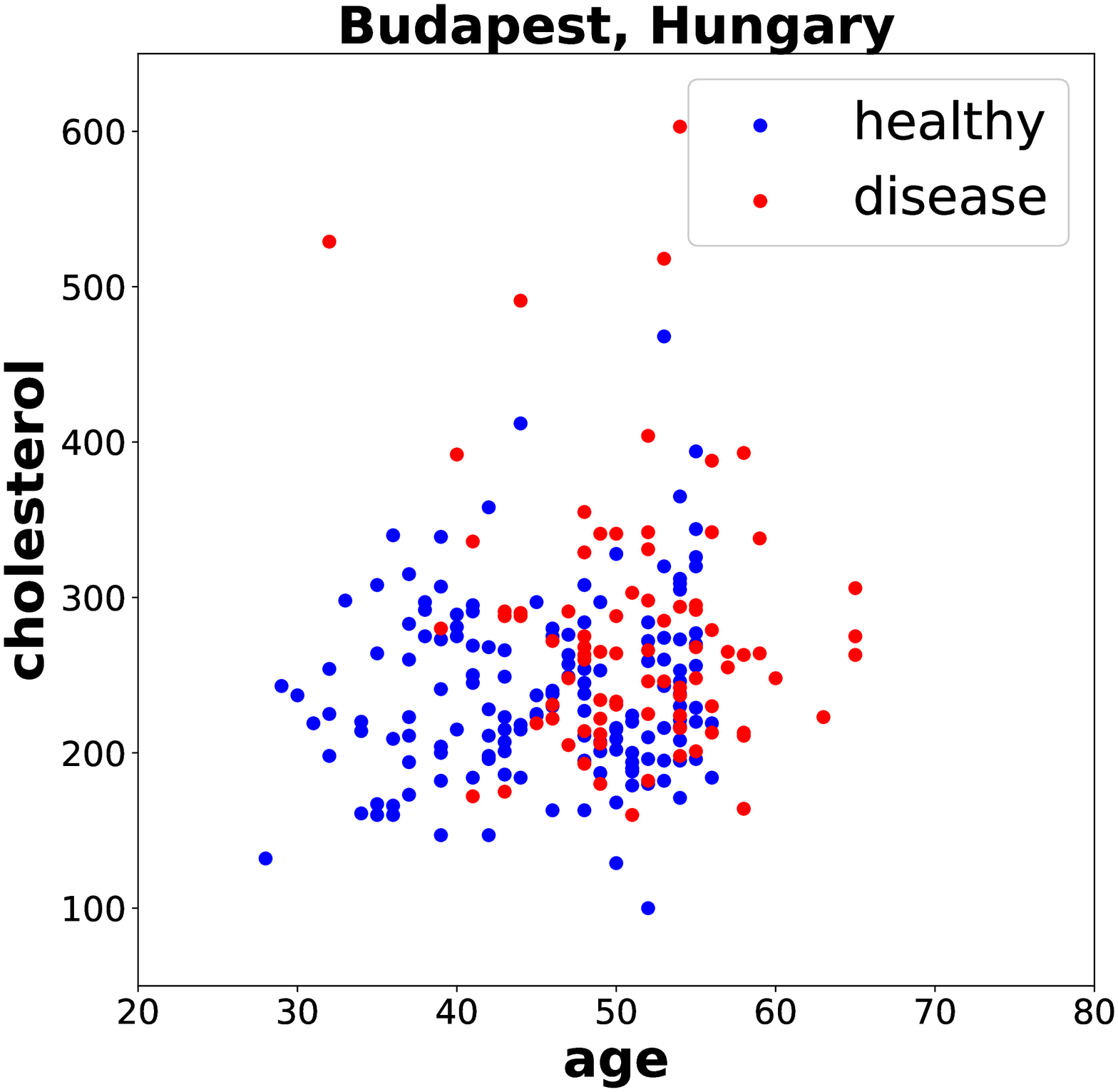} \
\includegraphics[height=120px]{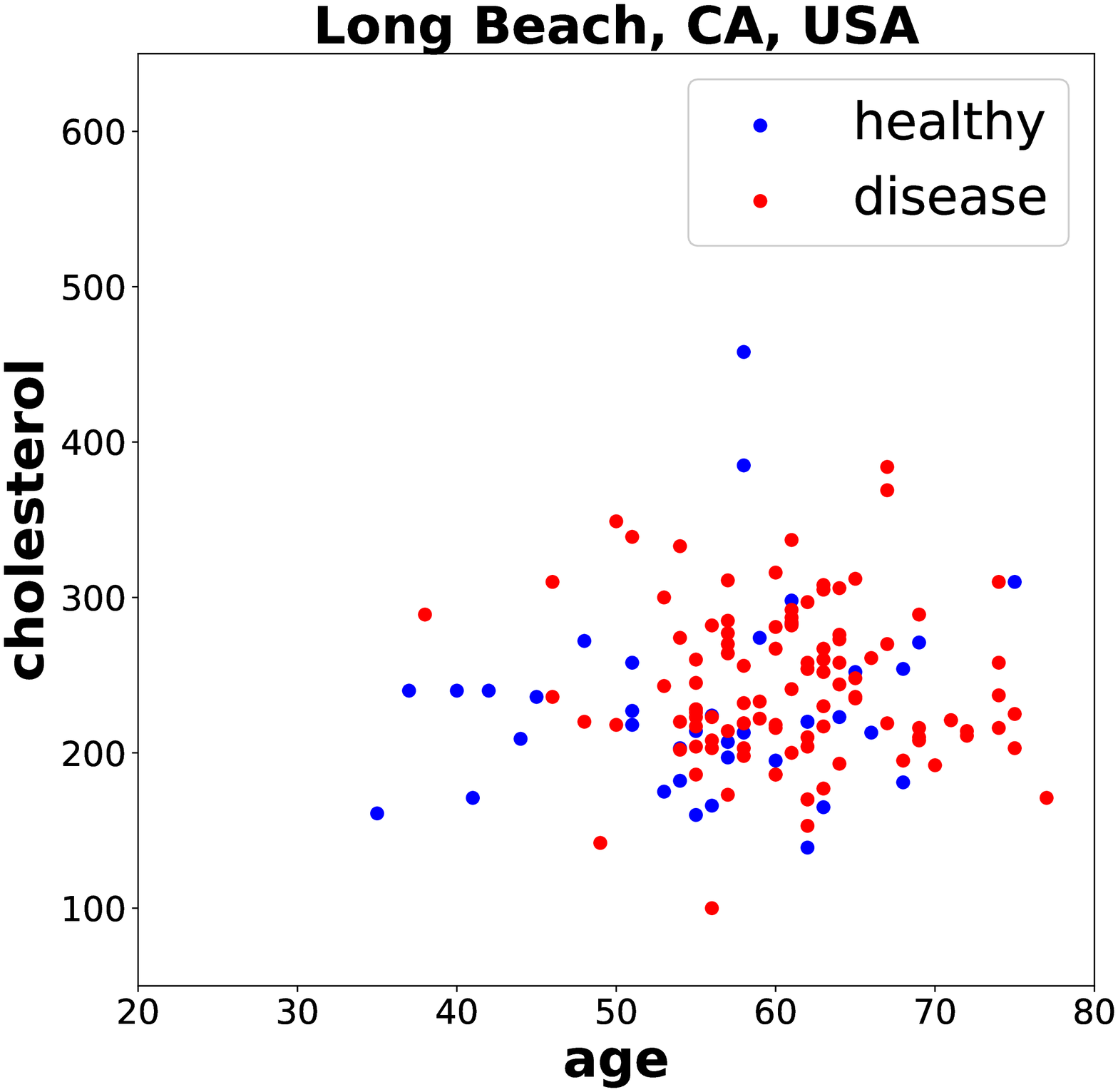}
\caption{Example of a domain adaptation problem, in which patients are diagnosed with heart disease based on their age and cholesterol. (Left) Data from the source domain, a hospital in Budapest. (Right) Data from the target domain, a hospital in Long Beach, California.}
\label{fig:example}
\end{figure}

We use a running example to illustrate the problem setting, and later on, to illustrate the behavior of some domain-adaptive classifiers. Figure \ref{fig:example} shows a scatter plot of measurements of patients from a hospital in Budapest, Hungary (left; the source domain) and patients from a hospital in Long Beach, California (right; the target domain). This a subset of the Heart Disease data set from UCI machine learning repository \cite{dua2019uciml}. The measurements consist of the patients' age (x-axis) and cholesterol level (y-axis) and the task is to predict whether they will develop heart disease (blue = healthy, red = heart disease).

\begin{figure}[thb]
\includegraphics[height=120px]{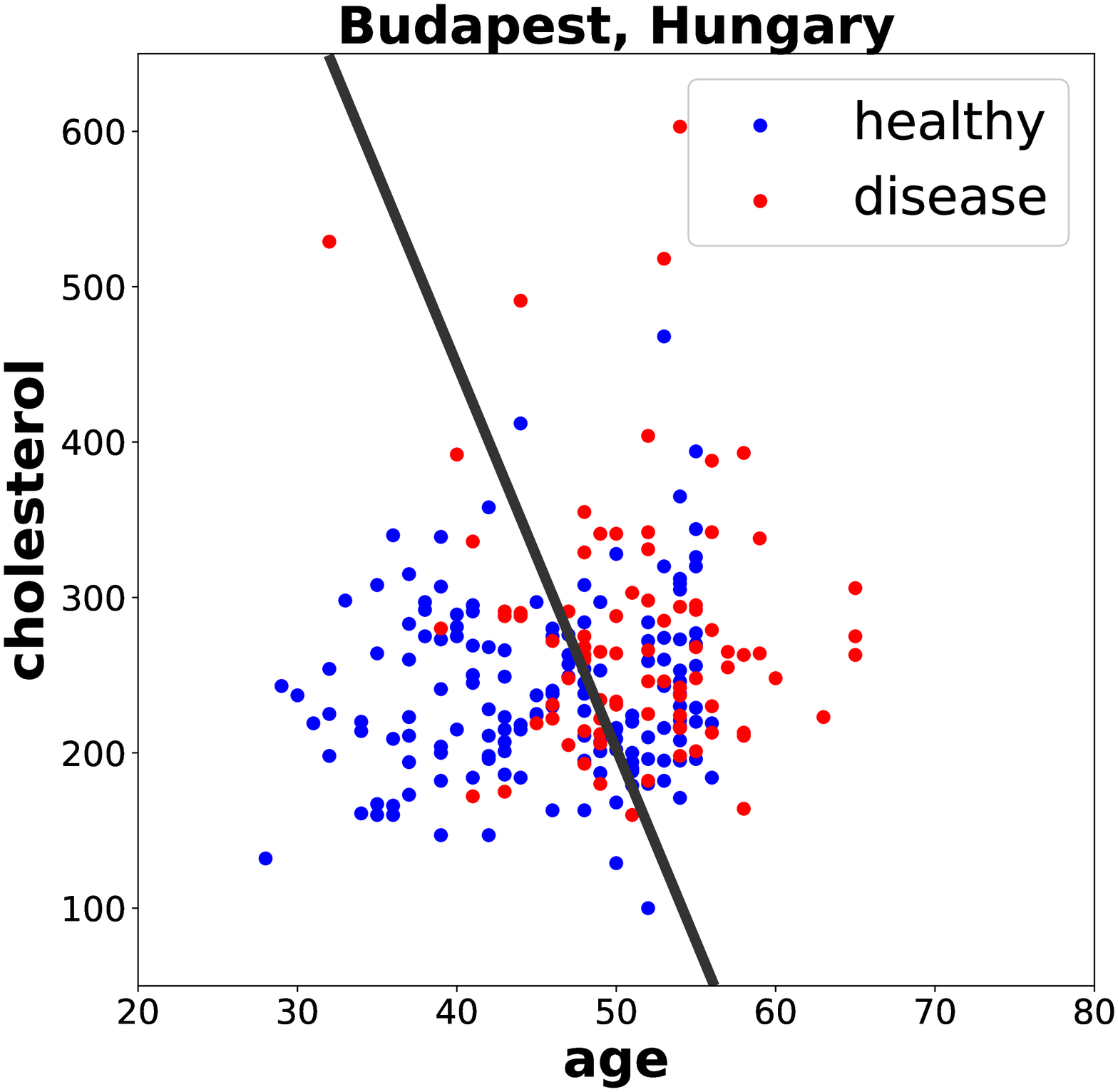} \
\includegraphics[height=120px]{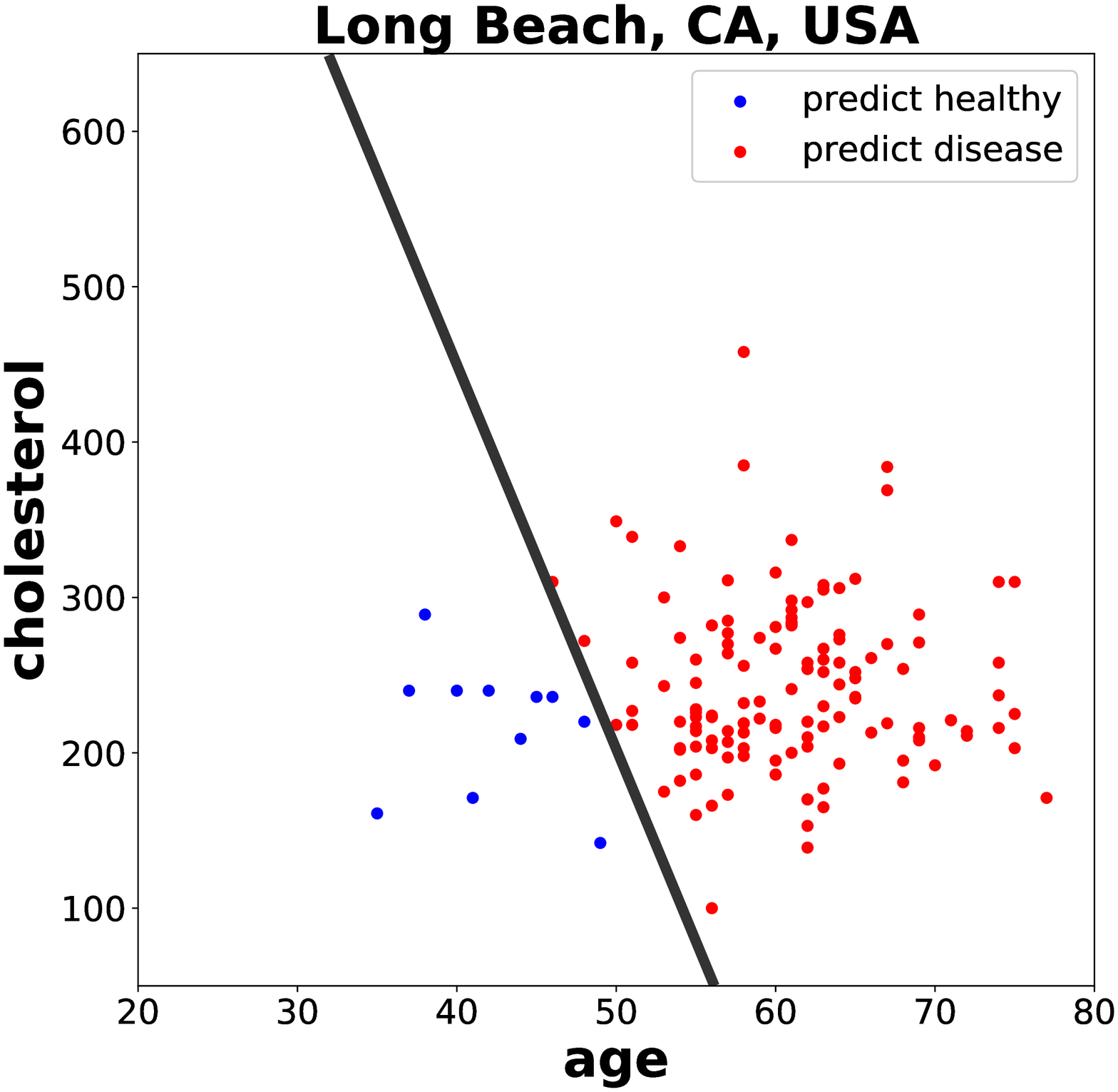}
\caption{Linear classifier (solid black line) trained on source data (left) and applied to target data (right).}
\label{fig:classifier}
\end{figure}

Figure \ref{fig:classifier} (left) shows the decision boundary of a linear classifier trained on the source samples (solid black line). 
It is not suited well to classifying samples from the target domain (Figure \ref{fig:classifier} right), which are shifted in terms of age. As can be imagined, its performance would decrease as the difference between the domains increases.
 
\subsection{Domain dissimilarity metrics} \label{sec:disc}
In order to characterize generalization across domains, a measure of domain dissimilarity is necessary. Many metrics of of differences between probability distributions or data sets exist, such as the Kullback-Leibler divergence, the total variation distance, the Wasserstein metric or the Kolmogorov-Smirnoff statistic \cite{mahmud2009universal,cover2012elements}. The choice of metric will often affect the behavior of a domain-adaptive classifier. We will discuss two measures in more detail, as they will appear later on in the paper. 
 
The \emph{symmetric difference hypothesis} divergence ($\mathrm{D}_{\mathcal{ H}\Delta \mathcal{ H}}$) takes two classifiers and looks at to what extent they disagree with each other on both domains \cite{ben2010theory}:
\begin{align}
	&\mathrm{D}_{\mathcal{ H} \Delta \mathcal{ H}} [p_{\mathcal{ S}}, p_{\mathcal{ T}}] = \nonumber \\
	&\ 2 \underset{h,h' \in \mathcal{ H}}{\sup} \ \left|  \ \text{Pr}_{\mathcal{ S}} \left[h(x) \neq h'(x) \right] - \text{Pr}_{\mathcal{ T}} \left[h(x) \neq h'(x) \right] \ \right|  \, . \label{sec:D_HDH}
\end{align}
In this context, it finds the pair of classifiers $h,h'$ for which the difference in disagreements between the source and target domain is largest \cite{ben2007analysis,ben2010theory}. Its value increases as the domains become more dissimilar.
Another example is the R{\'e}nyi divergence \cite{van2014renyi}:
\begin{align}
	\mathrm{D}_{\cal R^\alpha}[ p_{\cal T}, p_{\cal S}] = \frac{1}{\alpha - 1} \log_2 \int_{\cal X} p^{\alpha}_{\cal T}(x) / p^{\alpha-1}_{\cal S}(x) \mathrm{d}x \ \, , \label{eq:D_Renyi}
\end{align} 
where $\alpha$ denotes its order. For $\alpha=1$, the R{\'e}nyi divergence equals the Kullback-Leibler divergence \cite{van2014renyi}. In words, it corresponds to the expected value, with respect to the source distribution, of a power of the ratio of data distributions in each domain \cite{mansour2009multiple,cortes2010learning}. 

\subsection{Generalization error} \label{sec:geb}
Before one starts designing algorithms, one should consider whether it is at all possible to generalize across probability distributions. Chiefly, such questions are studied by examining the difference between the true error of an estimated classifier $\hat{h}$ and the true error of the optimal classifier $h^{*}$, known as the generalization error \cite{mohri2012foundations}. Bounding the generalization error often leads to insights on what conditions have to be satisfied to achieve certain levels of performance. 

First, we consider a generalization error bound for domain adaptation that does not incorporate an adaptation strategy. Instead, it characterizes the capacity of the source classifier to perform on the target domain in terms of domain dissimilarity, sample size and classifier complexity. Specifically, it relies on the error of the ideal joint hypothesis: $e^{*}_{\cal S,T} = \min_{h \in \mathcal{H}} \ [e_{\mathcal{ S}}(h) + e_{\mathcal{ T}}(h)]$ 
\cite{ben2007analysis,blitzer2008learning,ben2010theory}. 
If the error of the ideal joint hypothesis is too large, then there are no guarantees that the source classifier will perform well in the target domain. Given $e^{*}_{\cal S,T}$ and the symmetric difference hypothesis divergence $\mathrm{D}_{\mathcal{ H}\Delta \mathcal{ H}}$, one can state that with probability $1-\delta$, for $\delta \in (0,1)$, the following holds (Theorem 3 in \cite{ben2010theory} for $\beta=0$ and $\alpha=0$):
\begin{align}
	 e_{\mathcal{ T}}(\hat{h}_{\mathcal{ S}}&) - e_{\mathcal{ T}}(h_{\mathcal{ T}}^{*}) \ \leq 2 e^{*}_{\cal S,T} + \nonumber \\
	 &  \mathrm{D}_{\mathcal{ H}\Delta \mathcal{H}}(p_{\mathcal{ S}}, p_{\mathcal{ T}}) + 4 \sqrt{\frac{2}{n} \big(\nu \log (2 (n \! + \! 1)) \! + \! \log \frac{8}{\delta} \big)}  \, , 
\end{align}
where $\nu$ is the VC-dimension of the hypothesis space ${\cal H}$ \cite{mohri2012foundations}. In words: the generalization error of the source classifier with respect to the target domain is upper bounded by ideal joint hypothesis error, the domain dissimilarity and a factor consisting of classifier complexity and sample size. This bound is specific to error functions, but can be generalized to bounded real-valued loss functions \cite{mansour2009domain}. 

The above bound is loose because it does not incorporate an adaptation strategy. Now, the challenge is to find conditions that permit tighter bounds. In the methods we review, we have found assumptions that lead to bounds for adaptive classifiers. Firstly, if one assumes that the posterior distributions are equal in both domains $p_{\cal S}(y \given x) = p_{\cal T}(y \given x)$ (see Section \ref{sec:iw}), then the difference between the target errors of the optimal target classifier and an importance-weighted classifier can be bounded in terms of the R{\'e}nyi divergence between domains, the source sample size and classifier complexity \cite{cortes2010learning,azizzadenesheli2019regularized}. Secondly, if one assumes that there exists a set of components, obtainable through a non-trivial transformation of the features $t(x)$, that matches the class-conditional distributions $p_{\cal S}(t(x) \given y) = p_{\cal T}(t(x) \given y)$ (see Section \ref{sec:dic}), then one can bound the generalization error of a classifier trained on the transformed source data in terms of the divergence between the target data and the transformed source data \cite{gong2016domain}. Thirdly, assuming that a robust algorithm can achieve limited variation in loss for each partition it creates (see Section \ref{sec:robust}), then one can bound the difference between the target error and the average maximal loss, with respect to a shift in the posterior distributions, weighted by the probability of a sample falling in each partition \cite{mansour2014robust}. 
Lastly, if the assumption is made that the risk is small in the part of the target domain where the source domain is uninformative (see Section \ref{sec:PAC-Bayes}), then one can bound the difference between the source error weighted by the R{\'e}nyi divergence and the target error, of the Gibbs classifier \cite{germain2016new}. 

These assumptions provide some theoretical understanding of the domain adaptation problem, and more might exist still. They are insightful in that they reveal that generalization depends strongly on the particular form of domain dissimilarity. We will discuss them in more detail in their respective method categories.

\section{Sample-based approaches} \label{sec:sample}
In sample-based approaches, we are interested in minimizing the target risk through data from the source domain. One way to relate the source distribution to the target risk $R_{\cal T}$, at least superficially, is to consider:
\begin{align}
	R_{\mathcal{ T}}(h) =& \sum_{y \in Y} \int_{\mathcal{ X}} \ell(h(x), y) \ p_{\mathcal{ T}}(x,y) \ \mathrm{d} x \nonumber \\
	=& \sum_{y \in Y} \int_{\mathcal{ X}} \ell(h(x), y) \ \frac{p_{\mathcal{ T}}(x,y)}{p_{\mathcal{ S}}(x,y)} \ p_{\mathcal{ S}}(x,y) \ \mathrm{d} x \, . \label{eq:imprisk1}
\end{align}
In order to deal with Equation \ref{eq:imprisk1}, we need to determine the ratio $p_{\cal T}(x,y) / p_{\cal S}(x,y)$. Estimating that ratio would require labeled data from both domains. However, as discussed, target labels are considered unavailable. We therefore have to make simplifying assumptions so that risk estimation becomes possible without target labels.
 
Here we consider constrained forms of domain shifts: \emph{prior}, \emph{covariate} and \emph{concept} shift \cite{moreno2012unifying,storkey2009training}. Concept shift, also known as concept drift or conditional shift, will require observations of labeled data in both domains and is therefore out of our scope \cite{widmer1996learning}. Covariate shift corresponds to decomposing the joint distributions into $p(y \given x) p(x)$ and assuming that the posteriors remain equal in both domains, $p_{\cal S}(y \given x) = p_{\cal T}(y \given x)$ \cite{moreno2012unifying}. Conversely, prior shift, also referred to as label or target shift, corresponds to decomposing the joints into $p(x \given y) p(y)$ and assuming the conditional distributions remain equal $p_{\cal S}(x \given y) = p_{\cal T}(x \given y)$ \cite{moreno2012unifying}. It has been remarked in the causality community that covariate shift corresponds to causal learning (predicting effects from causes) and prior shift corresponds to anti-causal learning (predicting causes from effects) \cite{scholkopf2012causal}. 

Of course, in practice, the underlying probability distributions are unknown and assumptions on posteriors or conditionals cannot be verified. So, \emph{when} are the posterior or conditional distributions equal? These assumptions are known to hold in cases of sample selection bias \cite{heckman1977sample,storkey2009training}. In the sample selection bias setting, one differentiates between a true underlying data-generating distribution and a sampling distribution \cite{storkey2009training}. 
If the sampling distribution is not uniform, then the resulting data will be biased with respect to the underlying distribution. 
Since the underlying data-generating distribution remains constant, the underlying posteriors and conditionals remain equivalent between the biased and unbiased samples \cite{cortes2014domain}. Note that there are subtle differences between sample selection bias, covariate shift and domain adaptation: sample selection bias is a special case of covariate shift because it involves a sampling distribution and covariate shift is a special case of domain adaptation because the posteriors are assumed to be equal.

\subsection{Data importance-weighting} \label{sec:iw}
The knowledge that the posteriors remain equivalent can be exploited by decomposing the joint distributions and canceling terms:
\begin{align}
	R_{ W}(h) =& \sum_{y \in Y}\! \int_{\mathcal{ X}} \ell(h(x), y) \frac{\cancel{p_{\mathcal{ T}}(y \given x)} p_{\mathcal{ T}}(x)}{\cancel{p_{\mathcal{ S}}(y \given x)} p_{\mathcal{ S}}(x)} p_{\mathcal{ S}}(x,y) \mathrm{d} x \, . \label{eq:imprisk}
\end{align}
In this case, the importance weights consist of the ratio of the data marginal distributions, $w(x) = p_{\cal T}(x) / p_{\cal S}(x)$. 
A large weight indicates that the sample has high probability under the target distribution, but low probability under the source domain. As such, it is considered more important to the target domain than samples with small weights. 
The weights influence a classification model by increasing the loss for certain samples and decreasing the loss for others. Importance weighting is most often used in applications involving clinical or social science, but has been applied to natural language processing as well \cite{jiang2007instance}.

\begin{figure}[thb]
\includegraphics[height=120px]{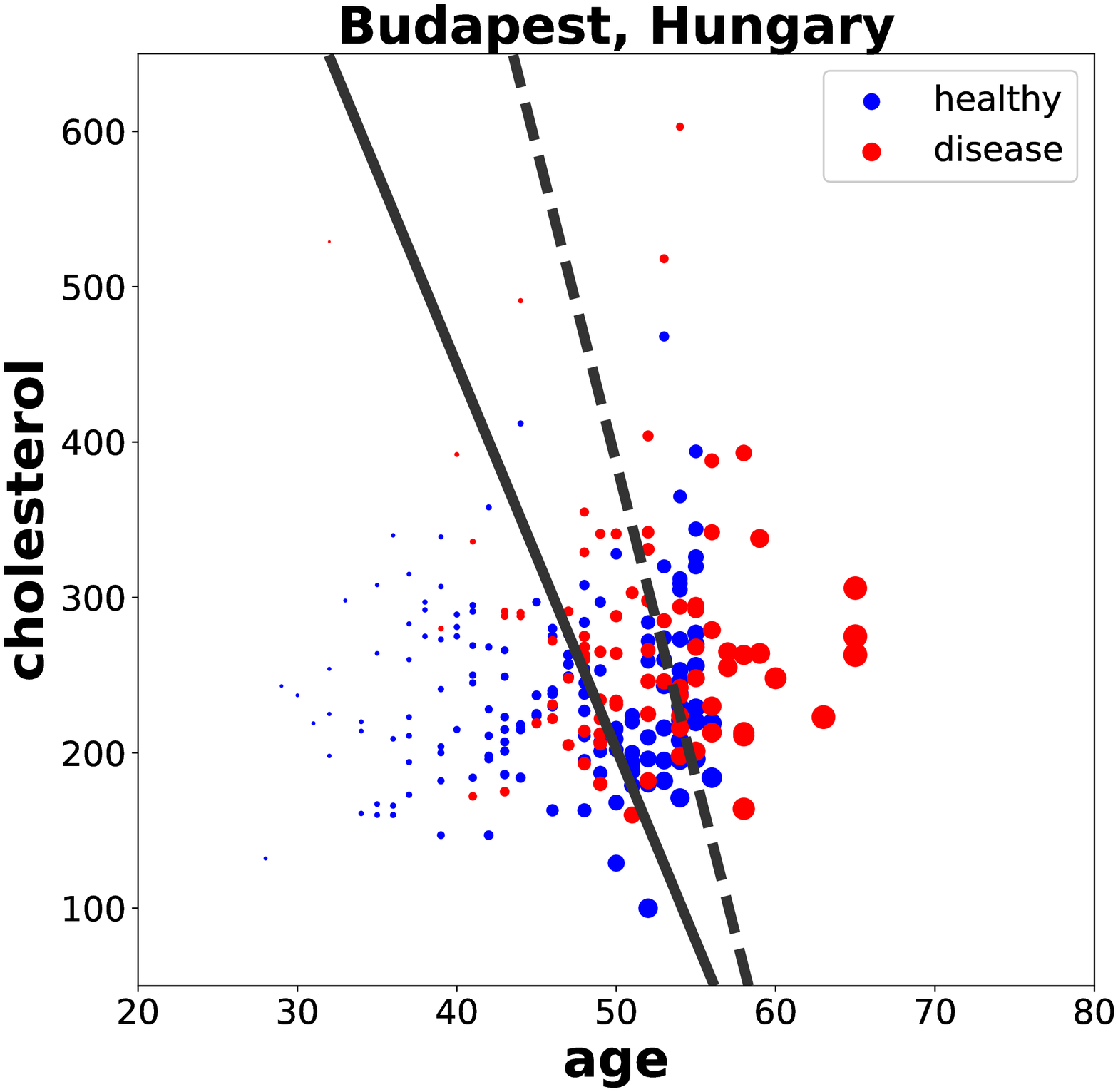} \
\includegraphics[height=120px]{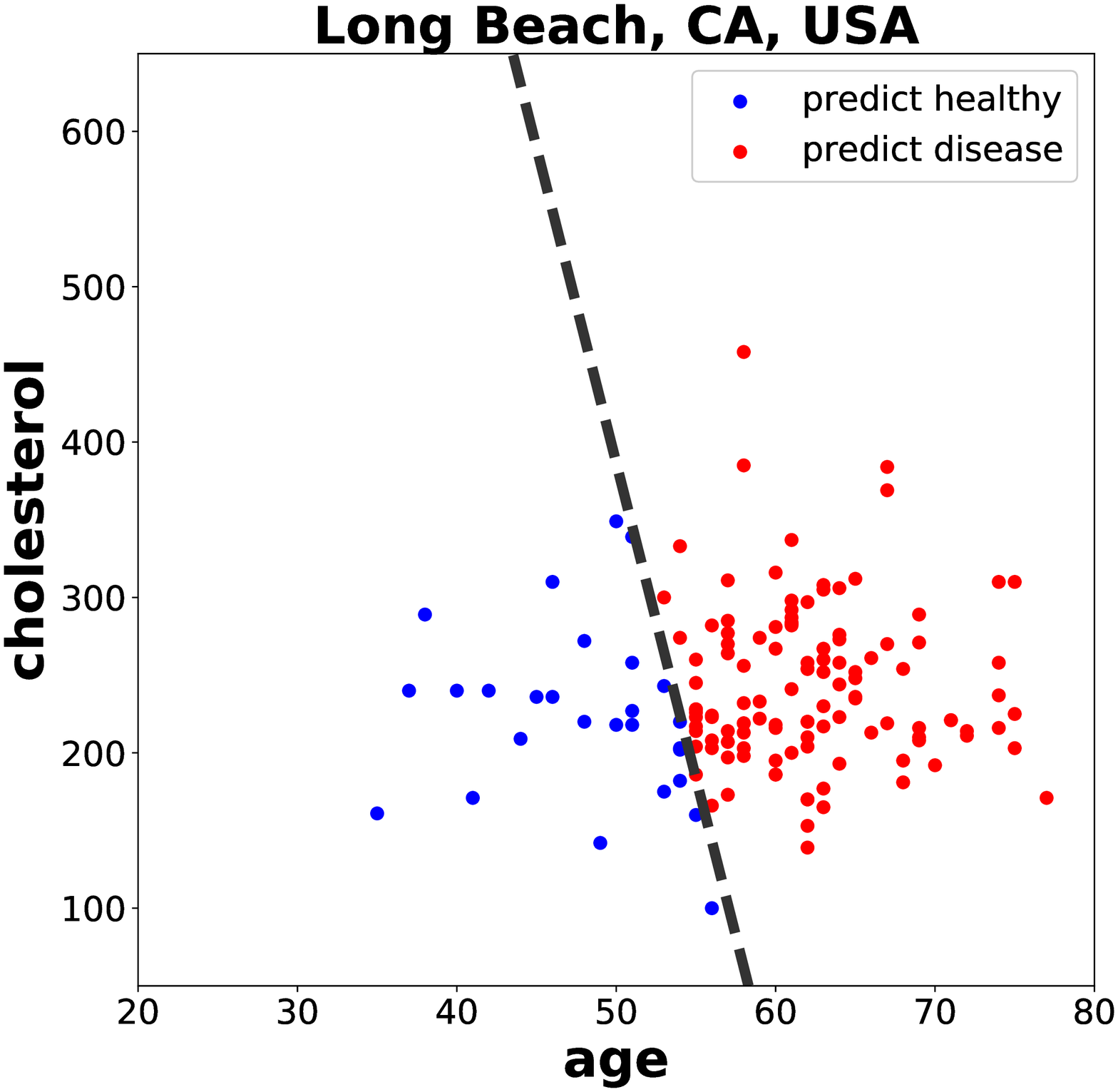}
\caption{Example of importance-weighting. (Left) The source samples from Figure \ref{fig:example} have been weighted (larger dot size is larger weight) based on their relative importance to the target domain, producing the importance-weighted classifier (black dashed line). (Right) Predictions made by the adapted classifier.}
\label{fig:iw}
\end{figure}

Figure \ref{fig:iw} shows a scatterplot of the example setting from Section \ref{sec:example}. The importance of the source samples is indicated with marker size. Training on these importance-weighted samples produces a decision boundary (dashed black line) that is different from the one belonging to the source classifier (solid black line) (see Figure \ref{fig:classifier}). The importance-weighted classifier is said to have \emph{adapted} to the data from the hospital in Long Beach. 

Equation \ref{eq:imprisk} is interesting, but does not tell us anything about the finite sample size case. For that, we will have a look at the difference between the true target error $e_{\cal T}$ and the empirical weighted source error $\hat{e}_{\cal W}$ of a given classifier $h$. With probability $1-\delta$ for $\delta > 0$ (Theorem 3, \cite{cortes2010learning}):
\begin{align}
e_{\cal T}(& h) - \hat{e}_{\cal W}(h) \nonumber \\
&\leq \ 2^{5/4} \ 2^{\mathrm{D}_{{\cal R}^2}[ p_{\cal T}, p_{\cal S}] / 2} \ \sqrt[3/8]{\frac{c}{n} \log\frac{2 n e}{c} + \frac{1}{n} \log \frac{4}{\delta}} \, , \label{sec:geb_iw}
\end{align}
where $\mathrm{D}_{{\cal R}^2}[ p_{\cal T}, p_{\cal S}]$ is the 2-order R{\'e}nyi divergence \cite{cortes2010learning}, and $c$ is the so-called \emph{pseudo-dimension} of the hypothesis space \cite{vidyasagar2002theory}. This bound indirectly tells us how far an importance-weighted classifier is from the optimal target classifier. Moreover, as the divergence between the domains increases, the sample size needs to increase at a certain rate as well, in order to maintain the same difference in error. Analysis of importance-weighting further reveals that an importance-weighted classifier will only converge if the expected squared weight is finite, i.e. $\mathbb{E}_{\cal S}[w(x)^{2}] < \infty$ \cite{cortes2010learning}. If the domains are too dissimilar, then this will not be the case. In addition, asymptotically, weighting is only effective for a \emph{mis-specified} classification model, e.g. a linear classifier for a non-linear classification problem \cite{white1981consequences,shimodaira2000improving,wen2014robust}. With a correctly specified model, as the sample size goes to infinity, the unweighted estimator will find the optimal classifier as well \cite{white1981consequences}. In fact, with a correctly specified model, the weighted estimator will converge to the optimal classifier for \emph{any} fixed set of non-negative weights that sums to $1$ \cite{white1981consequences}.

Given that we have mis-specified our model, how should we find appropriate importance weights? Indirect weight estimators first estimate the marginal data distribution of each domain separately and subsequently compute the ratio. Estimating each data distribution can be done parametrically, i.e. using a probability distribution with a fixed set of parameters, or non-parametrically, i.e. using a distribution with a variable set of parameters. With parametric weight estimators, one obtains a functional form of the resulting ratio of distributions. For example, one could assume Gaussian distributions for each domain \cite{shimodaira2000improving}. Then the weight function consists of:
\begin{align}
	\hat{w}(x_i) =& \ \frac{\mathcal{N}(x_i \given \hat{\mu}_{\cal T}, \hat{\Sigma}_{\mathcal{ T}})}{\mathcal{N}(x_i \given \hat{\mu}_{\cal S}, \hat{\Sigma}_{\cal S})} \, . 
\end{align}
Such weight functions can be analyzed and often show interesting properties. For example, should you choose members of the exponential family to estimate the data distributions, then you can expect the variance of the importance weights to increase drastically \cite{cortes2008sample,cortes2010learning}. High weight variance means that it is probable that a few samples will receive large weights. Consequently, at training time, the classifier will focus on those few samples designated as important, and ignore the others. The result is often a pathological classifier that will not generalize well.

The non-parametric alternative is to use kernel density estimators \cite{sugiyama2005input,yu2012analysis,baktashmotlagh2014domain}: 
\begin{align}
	\hat{w}(x_i) =& \frac{m^{-1}\sum_{j=1}^{m} \kappa_{\sigma_{\cal T}}(x_i - z_j)}{n^{-1} \sum_{i'=1}^{n} \kappa_{\sigma_{\cal S}}(x_i - x_{i'})}  \, ,
\end{align}
where $\kappa$ is a kernel function and $\sigma$ denotes its bandwidth. The density of a source sample $x_i$ depends on the distance to each kernel's center. Bandwidths are considered hyperparameters and can be tuned to produce smoother densities. Kernel density estimators have the advantage that they can become multi-modal: when samples are clustered in two regions, the overlayed kernel functions naturally form a mode on the cluster centers. 


Instead of estimating the data distributions in each domain and taking their ratio, the weights can be viewed as a separate variable \cite{sugiyama2008direct,sugiyama2012density}. Following this interpretation, the weights are directly estimated using an optimization procedure where the discrepancy between the weighted source and the target distribution is minimized \cite{tsuboi2009direct}. Different methods use different discrepancy measures. Two constraints are added to the optimization procedure. Firstly, all importance weights should be non-negative. Secondly, the weighted source distribution should be a valid probability distribution:
\begin{align}
    1 = \int_{\mathcal{ X}} p_{\mathcal{ T}}(x) \mathrm{d} x = \int_{\mathcal{ X}} w(x) p_{\mathcal{ S}}(x) \mathrm{d} x \approx \frac{1}{n} \sum_{i=1}^{n} w(x_i) \, . \label{eq:expweights}
\end{align} 
The above approximate equality can be enforced by constraining the absolute deviation of the weight average from $1$ to be less than some small $\epsilon$. With these two constraints, the optimization problem becomes: 
\begin{align}
	\hat{w} = \underset{w \in \mathbb{R}_{+}^{n}}{\arg \min} \ \mathrm{D} \big[ w, p_{\cal S}(x), p_{\cal T}(x) \big] 
\end{align}
such that 
$| n^{-1} \sum_{i=1}^{n} w_i - 1 | \leq \epsilon$.
$\mathrm{D}$ refers to the discrepancy measure, of which we discuss a few below. An advantage is that additional constraints can be incorporated to achieve certain desired effects, such as low weight variance. A limitation is that one needs to re-estimate weights when new source samples are added as it produces no weight function $w( \cdot )$, although some methods avoid this limitation \cite{sugiyama2005input,kanamori2009least}.

The most popular measure of domain dissimilarity is the Maximum Mean Discrepancy (MMD) \cite{borgwardt2006integrating,gretton2007kernel}. 
It is based on the distance between the expected values of two distributions under the continuous function that pulls them maximally apart \cite{borgwardt2006integrating}. But it can be approximated using universal kernels \cite{gretton2007kernel}. Minimizing the empirical MMD with respect to the importance weights, is called Kernel Mean Matching (KMM) \cite{huang2007correcting,gretton2009covariate}:
\begin{align}
	\mathrm{D}&_{\text{MMD}} \big[w, p_{\cal S}(x), p_{\cal T}(x) \big] = \Vert \ \mathbb{E}_{\cal S}[w \phi(x)] - \mathbb{E}_{\cal T}[ \phi(x)] \ \Vert_{\cal H} \nonumber \\
	&\propto \frac{1}{n^2} \sum_{i,i'}^{n} w_i \kappa(x_i,x_{i'}) w_{i'}) \! - \! \frac{2}{mn} \sum_{i}^{n} \sum_{j}^{m} w_i \kappa(x_i, z_j) \, ,
\end{align}
where $\phi$ is the basis expansion associated with the Gaussian kernel. Constant terms are dropped as they are not relevant to the optimization procedure. Depending on how the weights are further constrained, algorithmic computational complexities and convergence criteria can be derived as well \cite{gretton2009covariate,yu2012analysis}. Optimization consists of a quadratic program in the number of samples, which means that KMM in this form does not scale well to large data sets.

Another popular direct importance-weight estimator is the Kullback-Leibler Importance Estimation Procedure (KLIEP) \cite{sugiyama2005model,sugiyama2007covariate}. 
The KL-divergence between the true target distribution and the importance-weighted source distribution can be simplified to:
\begin{align}
	\mathrm{D}_{\text{KL}} \big[ p_{\mathcal{ S}}(x) &, p_{\mathcal{ S}}(x) w(x) \big] \nonumber \\
	=& \int_{\mathcal{ X}} p_{\mathcal{ T}}(x) \log \frac{p_{\mathcal{ T}}(x)}{p_{\mathcal{ S}}(x)} \mathrm{d} x - \int_{\mathcal{ X}} p_{\mathcal{ T}}(x) \log w(x) \mathrm{d} x \nonumber \\
	\propto& \ - \frac{1}{m} \sum_{j}^{m} \log w(z_j) \, . \label{eq:kliep1}
\end{align}
Note that this formulation requires weighting target samples. New weights would have to be estimated for each new target sample. To avoid this, a functional model is proposed. It consists of the inner product of parameters $\alpha$ and basis functions $\phi$, i.e. $w(x) = \phi(x) \alpha$ \cite{sugiyama2005input}. The objective now simplifies to $m^{-1} \sum_{j}^{m} \log \phi(z_j) \alpha$, where $\alpha$ is not dependent on an individual sample. So, KLIEP can be applied to new target samples without additional weight estimation. 

A third discrepancy is the $L^2$-norm between the weights and the ratio of data distributions \cite{kanamori2009least,kanamori2012statistical}. The squared difference can be expanded and terms not involving the weights can be dropped. Using a functional model of the weights, $w(x) = \phi(x) \alpha$, and approximating expectations with sample averages, the empirical discrepancy becomes:
\begin{align}
\mathrm{D}_{\text{LS}}& \big[w, p_{\mathcal{ S}}(x), p_{\mathcal{ T}}(x) \big] \nonumber \\
=& \ \frac{1}{2} \int_{\mathcal{ X}} \left(w(x) - \frac{p_{\mathcal{ T}}(x)}{p_{\mathcal{ S}}(x)} \right)^2 p_{\mathcal{ S}}(x) \mathrm{d} x \nonumber \\
\propto& \ \frac{1}{2} \ \alpha^{\top} \Big(\frac{1}{n} \sum_{i,i'}^{n} \phi(x_i)^{\top} \phi(x_{i'}) \Big) \alpha - \Big( \frac{1}{m} \sum_{j=1}^{m} \phi(z_j) \Big) \alpha \, , \label{LSIF}
\end{align} 
where $\phi(x_i)^{\top} \phi(x_{i'})$ denotes the outer product of applying the basis functions to a single source sample \cite{kanamori2009least}. This method is called the Least-Squares Importance Fitting procedure. Note that although this form resembles Kernel Mean Matching, it estimates $\alpha$'s, which can have a different dimensionality than the $w_i$'s. KMM can be very impractical in large data sets, because one needs to solve a quadratic program with $n$ variables, one for each $w_i$. Suppose the dimensionality of $\phi$ is $d$, and $d < n$. Then, there are $d$ $\alpha$'s, and $d$ variables for the quadratic program. As such, this approach can be computationally cheaper than kernel mean matching. It can also be extended to be kernel-based \cite{kanamori2012statistical}.

There are also direct weight estimators that do not employ optimization. A simple procedure is to use logistic regression to discriminate between samples from each domain \cite{sugiyama2012bregman}. In that case, the inverse estimated posterior probabilities become the importance weights. Additionally, Nearest-Neighbour Weighting is based on tessellating the feature space into Voronoi cells, which approximate a probability distribution function in the same way as a multi-dimensional histogram \cite{miller2003new,loog2012nearest}. To obtain weights, one forms the Voronoi cell $V_i$ of each source sample $x_i$ with the part of feature space that lies closest to $x_i$ \cite{loog2012nearest}. The ratio of target over source is then approximated by counting the number of target samples $z_j$ that lie within each Voronoi cell: 
\begin{align}
	\hat{w}_i = |V_i \cap \{z_j\}_{j=1}^{m} | \, ,
\end{align}
where $| \cdot |$ denotes cardinality. Counting can be done by a nearest-neighbours algorithm \cite{loog2012nearest,kremer2015nearest}. This estimator does not require hyperparameter optimization, but Laplace smoothing, which adds a 1 to each cell, can additionally be performed \cite{loog2012nearest}.

Lastly, it is also possible to avoid the two-step procedure, and optimize the weights simultaneously with optimizing the classifier \cite{bickel2007discriminative,bickel2009discriminative}.  

\subsection{Class importance-weighting}
Similar to before, the knowledge that the conditionals are equivalent can be exploited by canceling them. This produces the following weighted risk function for prior shift correction:
\begin{align}
	R_{ W}(h) \! =& \! \sum_{y \in Y}\! \int_{\mathcal{ X}} \! \ell(h(x), y) \frac{\cancel{p_{\mathcal{ T}}(x \given y)} p_{\mathcal{ T}}(y)}{\cancel{p_{\mathcal{ S}}(x \given y)} p_{\mathcal{ S}}(y)} p_{\mathcal{ S}}(x,y) \mathrm{d} x \, . \label{eq:prior_shift}
\end{align}
where $w(y) = p_{\cal T}(y) / p_{\cal S}(y)$ are the class weights, correcting for the change in class priors between the domains. Weighting in this manner is related to cost-sensitive learning \cite{elkan2001foundations} and class imbalance \cite{jacobusse2016selection}. The same effect can be achieved by under- or over-sampling data points from one class \cite{chawla2002smote}. 

Nonetheless, weighting based on target labels is outside the scope of this review, and we will not discuss it further. In contrast, there are a variety of methods that estimate class importance weights without requiring target labels \cite{zhang2013domain,lipton2018detecting,azizzadenesheli2019regularized}. In Black Box Shift Estimation (BBSE), one takes a black-box predictor $h$, computes the confusion matrix $\mathrm{C}_{h(x), y}$ on a validation split of the source data and makes predictions for the target data \cite{lipton2018detecting}. The class weights can then be obtained by taking the product of the inverse confusion matrix and the predicted target prior: 
\begin{align}
	\hat{w}_k = \mathrm{C}^{-1}_{h(x), k} \ \hat{p}_{\cal T}(h(z) = k) \, ,
\end{align}
where $\hat{p}_{\cal T}(h(z))$ are the empirical proportions of the classifier's predictions. The difference between the estimated and the true class weights can be bounded tightly \cite{lipton2018detecting}. The main difficulty in estimating class weights lies in the fact that estimating the confusion matrix becomes unstable if it is nearly singular, which is especially a problem when the sample size is small \cite{azizzadenesheli2019regularized}. In BBSE, this is avoided by using soft labels instead of hard ones \cite{lipton2018detecting}. Alternatively, one could regularize the weight estimator \cite{azizzadenesheli2019regularized}. Regularized Learning of Label Shift first reparameterizes the class weights to $\upsilon = w - 1$ and then estimates the reparametrized weights using a stable procedure. They are transformed back into the weights, $\hat{w} = 1 + \lambda \hat{\upsilon}$, using a regularization term $\lambda$ that depends on the source sample size. The procedure lends itself well to analysis and a generalization error bound on the class importance-weighted estimator can be formed (Theorem 1, \cite{azizzadenesheli2019regularized}). This bound also utilizes the R{\'e}nyi divergence, similar to the one for data importance-weighting \cite{cortes2010learning}.

Kernel Mean Matching has also been extended to estimating weights for prior shift \cite{zhang2013domain}. The class-conditional distributions in the source domain can be estimated and averaged over the weighted source priors to produce a new data marginal distribution. Using the Maximum Mean Discrepancy between the target data distribution and the new data marginal, the following class weight estimator is obtained:
\begin{align}
	\hat{w}_k \! = \! \underset{w \in \mathbb{R}_{+}^{K}}{\arg \min}\ \| U \left[p_{\cal S}(x \given y) \right] \mathbb{E}_{\cal S}[w \psi(y)] - \mathbb{E}_{\cal T}[\phi(x)] \|^2  \, ,
\end{align}
where $U\left[ p_{\cal S}(x \given y) \right]$ is the product of the cross-covariance $C_{\cal X,Y}$ and the inverse class covariance $C_{\cal Y,Y}^{-1}$, and $\psi(y)$ is a basis expansion of the weights. 
The prior shift assumption that the class-conditionals are equal can be even relaxed to allow for location-scale changes \cite{zhang2013domain}. 

\section{Feature-based approaches} \label{sec:feature}
\begin{figure}[thb]
\centering
\includegraphics[height=120px]{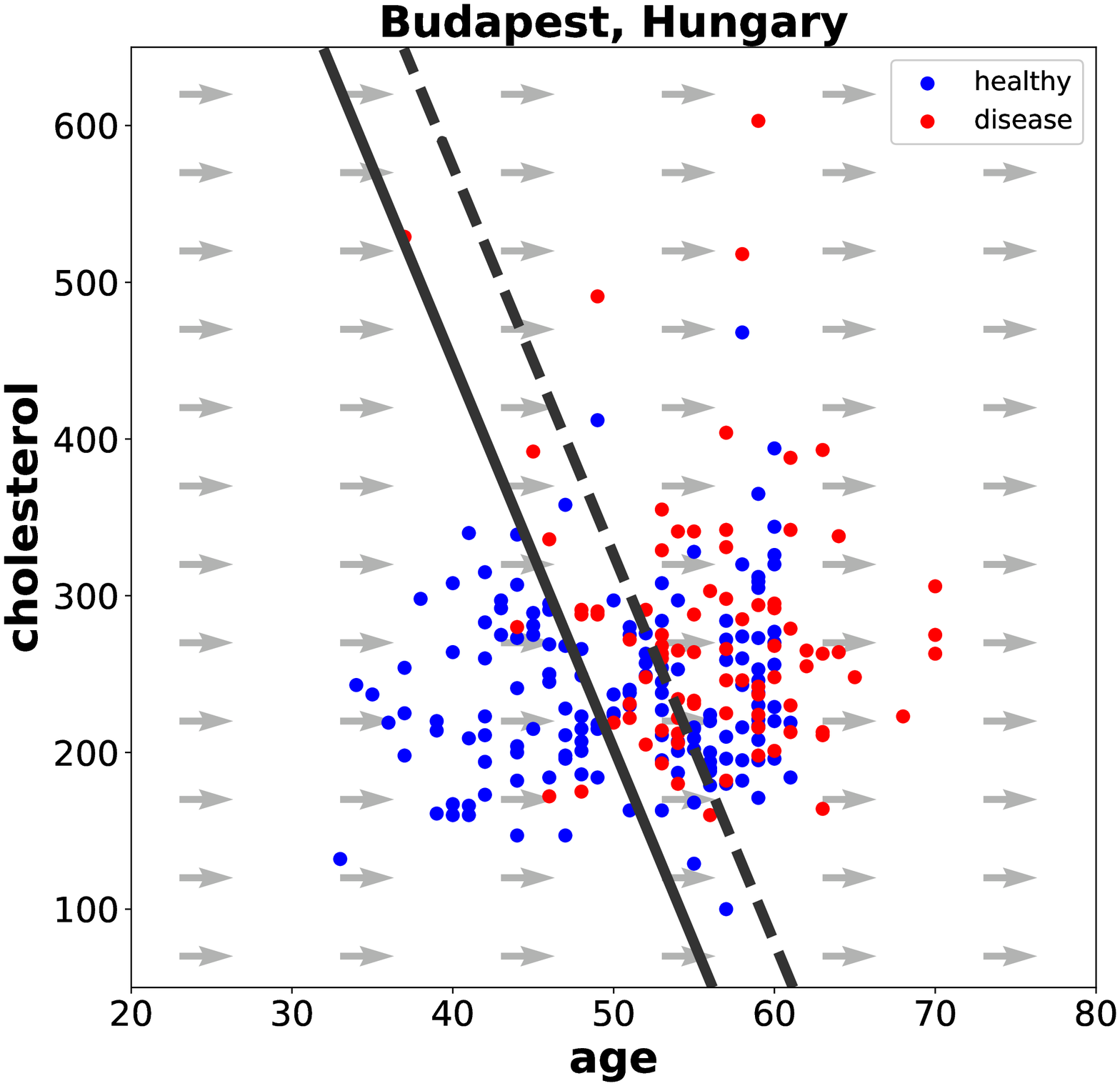}
\includegraphics[height=120px]{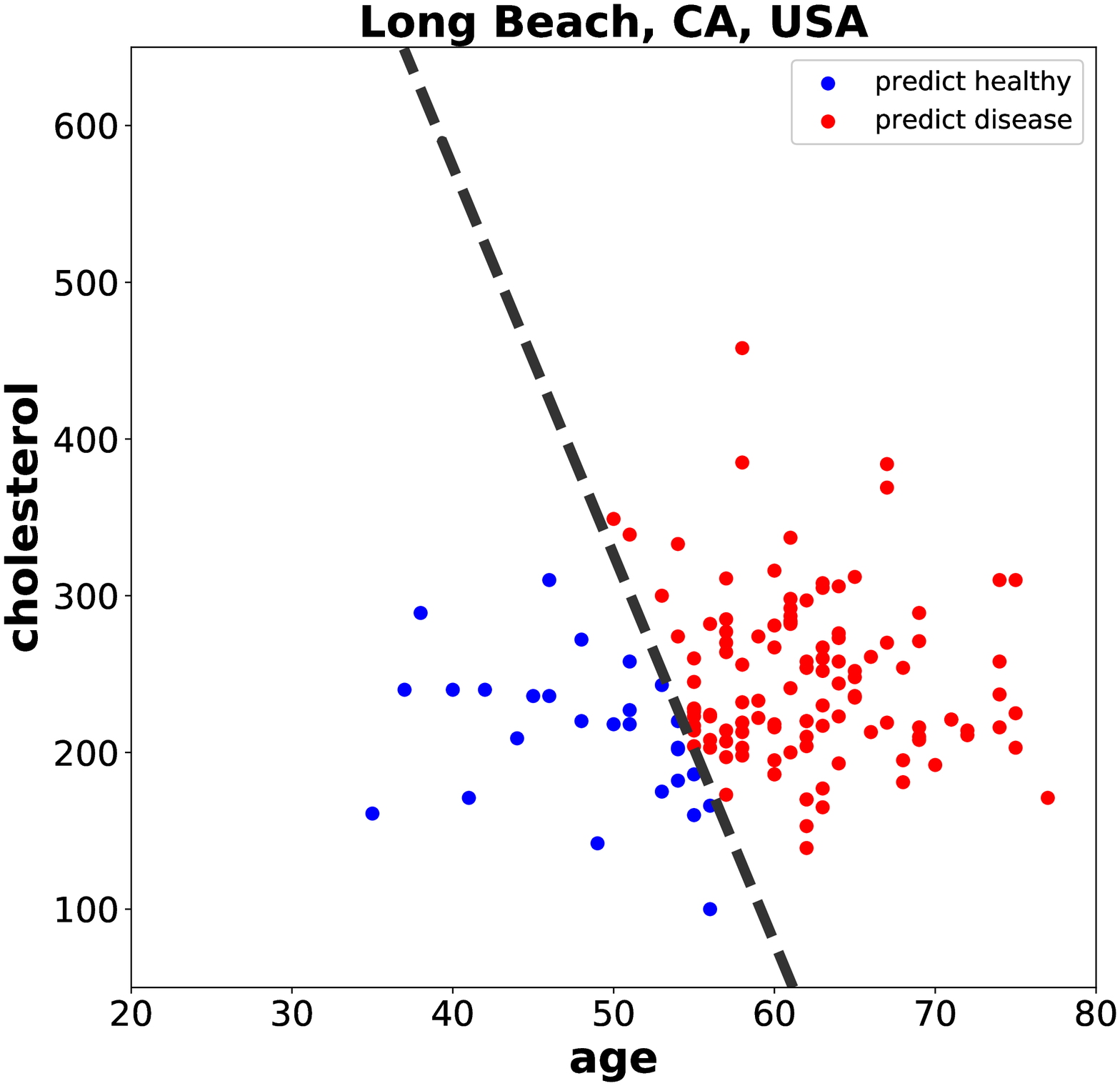}
\caption{Example of a feature-based method. (Left) The source samples from Figure \ref{fig:example} have been translated to match the data from the target domain. Subsequently, a classifier is trained on the mapped source data (black dashed line). The original naive classifier (black solid line) is shown for comparison. (Right) The adapted classifier is applied to the target samples and predictions are shown.}
\label{fig:subm}
\end{figure}

In some problem settings, 
there potentially exists a transformation that maps source data onto target data \cite{gopalan2015domain,patel2015visual}. In the example setting, the average age of patients in Budapest is lower than that of the patients in Long Beach. One could consider reducing the discrepancy between the hospitals by shifting the age of the Hungarian patients upwards and training a classifier on the shifted data. Figure \ref{fig:subm} (left) visualizes such a translation with the vector field in the background. The adapted decision boundary (black dashed line) is a translated version of the original naive source classifier (black solid line).

It has been argued that matching source and target data is justified by the generalization error bound from Section \ref{sec:geb} \cite{fernando2013unsupervised,long2015domain,tzeng2015simultaneous,ganin2016domain}. That would shrink the domain discrepancy and produce a tighter bound on the generalization error. However, note that matching the data distributions, $p_{\cal S}( t(x)) \approx p_{\cal T}(x)$ where $t(x)$ is the transformation function, does not imply that the conditional distributions will be matched as well, $p_{\cal S}(y \given t(x)) \approx p_{\cal T}(y \given x)$ \cite{zhang2013domain,gong2016domain}. To achieve such a result, stronger conditions are necessary (see Sections \ref{sec:opt_transport} and \ref{sec:dic}).

\subsection{Subspace mappings}
In certain problems, the domains contain domain-specific noise but common subspaces. Adaptation would consist of finding these subspaces and matching the source data to the target data along them. One of the most straightforward of such techniques, called Subspace Alignment, computes the first $d$ principal components in each domain, $C_{\mathcal{ S}}$ and $C_{\mathcal{ T}}$, where $d < D$ \cite{fernando2013unsupervised}. A linear transformation matrix is then computed that aligns source components to target components: $M = C_{\cal S}^{\top}C_{\cal T}$. The adaptive classifier projects data in each domain to their components, maps the projected source data using $M$ and trains on the transformed source data. Extensions of this approach include basing the alignment on particular landmarks (Landmarks Selection-based Subspace Alignment) \cite{aljundi2015landmarks}, basing the alignment on both the data distributions and the subspaces (Subspace Distribution Alignment) \cite{sun2015subspace} and training a classifier jointly with learning a subspace (Joint cross-domain Classification and Subspace Learning) \cite{fernando2015joint}.

A slightly more complicated technique is to model the structure of the data with graph-based methods, instead of aligning directions of variance \cite{das2018unsupervised,das2018graph}. Data is first summarized using a subset called the exemplars, obtained through clustering. From the set of exemplars, two hyper-graphs are constructed. These two hyper-graphs are matched along their first, second and third orders using tensor-based algorithms \cite{das2018unsupervised,das2018graph}. Higher-order moments consider other forms of geometric and structural information, beyond pairwise distances between exemplars \cite{das2018sample,das2018unsupervised}.

A further step can be taken by assuming that there exists a \emph{manifold} of transformations between the source and target domain \cite{gong2013reshaping,baktashmotlagh2014domain,hoffman2014continuous}. This manifold consists of a space of parameters, where each point would generate a possible domain. For example, the manifold might consist of a set of camera optics parameters, where each setting would produce data in a different part of feature space  \cite{gopalan2011domain,baktashmotlagh2014domain}. The manifold assumption is interesting, because it implies that there exists a path along the manifold's surface from the source to the target domain \cite{baktashmotlagh2013unsupervised,caseiro2015beyond}. Every point along that path could generate an intermediate domain \cite{gong2013connecting,gopalan2014unsupervised}. For example, the space of all linear subspace transformations is termed the \emph{Grassmann} manifold \cite{zheng2012grassmann}. A path along the Grassmannian generates many linear transformations, which could, in turn, generate intermediate domains \cite{gopalan2014unsupervised}. These intermediate domains can be used during training to inform a classifier on how to adapt its decision boundary.

But it is also possible to incorporate the entire path. Geodesic Flow Kernel forms a kernel consisting of the inner product between two feature vectors $x_i$ and $x_j$ projected onto the $t$-th subspace, for $t \in [0,1]$ \cite{gong2012geodesic}. It then integrates over $t$ to account for all intermediate subspaces: $G(x_i, x_j) = \int_{0}^{1} x_i M(t) M(t)^{\top} x_{j}^{\top} \ \mathrm{d}t$
where $M(t)$ is the projection matrix at time $t$. At $t=0$, the projection consists of purely the source components, $M(0) = C_{\cal S}$, and at $t=1$, it consists purely of the target components, $M(1) = C_{\cal T}$. The resulting kernel can be used in combination with a support vector machine or a kernel nearest-neighbour \cite{gong2012geodesic,gong2013connecting}.

An alternative is to consider statistical manifolds \cite{baktashmotlagh2014domain}. A path on the statistical manifold may describe a sequence of parameters that turns one distribution into another, for instance a sequence of means between two Gaussian distributions. The length of the geodesic path along the statistical manifold is called the Hellinger distance, and can be used as a measure of domain discrepancy \cite{ditzler2011hellinger}. Adaptation consists of finding a sequence of parameters based on minimum Hellinger distance, and use these in a similar fashion as the Geodesic Flow Kernel \cite{baktashmotlagh2014domain,baktashmotlagh2016distribution}.

Lastly, if one believes that the transformation from source to target is stochastic, one could consider a stochastic mapping. Feature-level domain adaptation (FLDA) minimizes risk under a transfer model describing the probability where a source sample would lie in feature space, if it were part of the target domain \cite{kouw2016feature}. For example, suppose that data is collected with a rich set of features in the source domain, but in the target domain, there is a chance that some features will not be collected. A distribution could be fitted to source and target data to estimate the probability of a feature dropping out. The FLDA classifier trained under this transfer model will effectively not rely on features with a high probability of dropping out \cite{kouw2016feature}.

\subsection{Optimal transport} \label{sec:opt_transport}
Another class of transformation-based techniques is optimal transport. In optimal transport, one estimates probability measures from data and finds a mapping of minimal cost that equates the measures \cite{courty2016optimal,courty2017joint}. 
For domain adaptation, one assumes that there exists a transformation $t(\cdot)$ that matches the source and target posterior distributions in the following way: $p_{\cal T}(y \given t(x)) = p_{\cal S}(y \given x)$ \cite{courty2016optimal}. The transportation map $t$ is also called a \emph{push-forward}, in this case from $p_{\cal S}$ to $p_{\cal T}$. After applying the transformation, the source samples resemble the target samples and a classifier trained on the transformed source samples would be suitable to classifying target samples.

Finding the optimal transportation map between two measures among the set of all possible transformations is intractable \cite{courty2016optimal}. Therefore, the optimization problem is relaxed into a search over a joint probability measure $\gamma$, with marginals $p_{\cal S}(x)$ and $p_{\cal T}(x)$, that has minimal distance between points. The joint measure couples the two marginals together, allowing one to be obtained from the other. This new objective corresponds to the Wasserstein distance:
\begin{align}
	\mathrm{D}_{\cal W}[p_{\cal S}(x), p_{\cal T}(x)] = \underset{\gamma \in \Gamma}{\inf} \int_{\cal X \times X} d(x, z) \mathrm{d}\gamma(x, z) \, ,
\end{align}
where $\Gamma$ refers to the set of all joint measures on ${\cal X} \times {\cal X}$ with marginals $p_{\cal S}(x)$ and $p_{\cal T}(x)$. The function $d(\cdot, \cdot)$ refers to a distance metric between two points, usually taken to be the Euclidean norm \cite{courty2016optimal}. The joint measure $\gamma^{*}$ that minimizes the Wasserstein distance between the two marginals is called the \emph{transportation plan}. 

In practice, the data marginals are estimated using the empirical distributions, $\hat{p}_{\cal S}(x) = \sum_{i=1}^{n} P^{\cal S}_i \delta(x - x_i)$ and $\hat{p}_{\cal T}(x) = \sum_{j=1}^{m} P^{\cal T}_j \delta(z - z_j)$ where $\delta(\cdot)$ is a Dirac function and $P_i$ is the probability mass assigned to the $i$-th sample \cite{redko2017theoretical}. The set $\Gamma$ becomes the set of all non-negative matrices of size $n \times m$ where the sum over columns corresponds to the empirical source distribution $\hat{p}_{\cal S}$ and the sum over rows corresponds to the empirical target distribution $\hat{p}_{\cal T}$ \cite{courty2016optimal}. The transportation plan $\gamma^{*}$ can now be found by minimizing the inner product of $\gamma$ and the distance matrix: $\gamma^{*} = \arg \min_\gamma \| \gamma^{\top} C \|_{F}$ where $\| \cdot \|_{F}$ denotes the Frobenius norm and $C_{ij} = d(x_i, z_j)$ is the distance matrix \cite{courty2016optimal}. This minimization is expensive due to its combinatorial nature, but it can be relaxed through regularization. To actually transform the source samples, one takes the barycentric mapping, $\tilde{x}_i = \arg \min_ {x} \sum_{j=1}^{m} \gamma^{*}_i d(x, z_j)$ \cite{courty2016optimal}. 



The advantage of the optimal transport formulation is that it is possible to exploit structure in the problem setting. For instance, if the source data conforms to a graph, that graph can directly be incorporated through Laplacian regularization \cite{courty2016optimal}. Additionally, the measures need not be data marginals, but could also be joint distrbutions \cite{courty2017joint}. Furthermore, generalization error bounds can be formed, with forms similar to the one described in Section \ref{sec:geb} \cite{courty2017joint,redko2017theoretical}. The Wasserstein distance is important to generative adversarial networks, and there have been a number of works combining optimal transport with deep learning \cite{bhushan2018deepjdot,chen2018re,le2018theoretical}. 

\subsection{Domain-invariant spaces} \label{sec:dic}
The problem with transforming data from one domain to match another, is that the representation remains domain-specific. But variation due to domains is often more of a nuisance than an interesting factor. Ideally, we would like to represent the data in a space which is domain-invariant. The advantage of this approach is that classification becomes the same as standard supervised learning. Most domain-invariant projection techniques stem from the computer vision and image processing communities, where domains are often caused by image acquisition procedures \cite{torralba2011unbiased}.


In cases where there is a causal relationship between the variables ${\cal X}$ and ${\cal Y}$ such that ${\cal Y}$ causes ${\cal X}$, it is possible to bound the difference between the class-conditional distributions in terms of the difference between the data distributions \cite{zhang2013domain,gong2016domain}. Under the conditions specified by the bound, mapping data to a domain-invariant space would also match class-conditional distributions. To explain this result, we must first define the notion of \emph{Conditional Invariant Components}: $X^{ci}$ are $d$-dimensional components of ${\cal X}$, obtained by transforming the data $X^{ci} = t(x)$, such that $p_{\cal S}(x^{ci} \given y) = p_{\cal T}(x^{ci} \given y)$ \cite{gong2016domain}. Now, it is necessary to make two assumptions. Firstly, the transformation $t$ is assumed to be non-trivial. A trivial transformation is one where the conditional distributions of the transformed data are dependent for each class. For example, multiplying each feature with $0$. Secondly, all linear combinations of the source class-conditional and the target class-conditional distributions are assumed to be linearly independent of each other, for any two classes. If these two assumptions hold, then the following holds (adapted from Lemma 1 and Theorem 2 from \cite{gong2016domain}, assuming equal class priors $p_{\cal S}(y) = p_{\cal T}(y)$):
\begin{align}
	e_{\cal T}(h) \! - \! e_{\tilde{\cal S}}(h) \leq J^{ci} \mathds{1}_{(0,\pi/2]}(\theta) \! + \! \frac{2}{\sin^2 \theta} J^{ci} \mathds{1}_{(\pi/2, \pi)}(\theta) , 
\end{align}
where $e_{\tilde{\cal S}}(h)$ is the error on the transformed data, $p_{\cal S}(t(x), y)$, and $\mathds{1}$ is the indicator function. $J^{ci} = \| \mathbb{E}_{\cal S}[ \phi(t(x))] - \mathbb{E}_{\cal T}[ \phi(t(x)) \|^2$ is the MMD divergence between the transformed source and the transformed target data. When $J^{ci}$ is $0$, the source distribution is transformed perfectly in the target distribution. The difference between the error on the transformed source data and the target error will then be $0$. Above, $\theta$ is the angle between two Kronecker delta kernels, where that kernel is: $	\Delta_{c} = \ p_{\cal S}(y = c) \ \mathbb{E}_{x^{ci} \sim p_{\cal S}(x^{ci} \given y = c)} \big[\phi(x^{ci}) \big] - p_{\cal T}(y = c) \ \mathbb{E}_{x^{ci} \sim p_{\cal T}(x^{ci} \given y = c)} \big[\phi(x^{ci}) \big]$ 
for $c \in {\cal Y}$. Each delta kernel consists of the difference between the source and target distributions over the invariant components for that class, weighted by the class prior. If $\theta$ is between $\pi/2$ and $\pi$, the bound becomes looser as $\theta$ goes to $\pi$. At $\theta = \pi$, $J^{ci}$ cannot be used as an upper bound.

A simple approach to finding a domain-invariant space is to find principal components based on joint directions of variation \cite{pan2010cross,pan2011domain}. In order to do so, the joint domain kernel, $\mathrm{K} = [\kappa_{\mathcal{ S,S}}  \ \kappa_{\mathcal{ S,T}}; \ \kappa_{\mathcal{ T,S}} \ \kappa_{\mathcal{ T,T}}]$, is first constructed \cite{pan2008transfer}. Data projected onto components $C$ should have minimal distance to the empirical means in each domain \cite{pan2011domain}. As such, components are extracted by minimizing the trace of the projected joint domain kernel:
\begin{align}
	\underset{C}{\text{minimize}} \quad &\text{tr}(C^{\top}\mathrm{K} \mathrm{L} \mathrm{K} C) \nonumber \\
	\text{s.t.} \quad & C^{\top}\mathrm{K} \mathrm{H} \mathrm{K}C = I \, ,
\end{align}
where $\mathrm{L}$ is the normalization matrix that divides each entry in the joint kernel by the sample size of the domain from which it originated, and $\mathrm{H}$ is the matrix that centers $\mathrm{K}$ \cite{pan2011domain}. The constraint is necessary to avoid trivial solutions, such as projecting all data to $0$. This technique is known as Transfer Component Analysis \cite{pan2011domain}. It resembles kernel PCA and, likewise, its optimization consists of an eigenvalue decomposition \cite{scholkopf1997kernel}. A regularization term $\text{tr}(C^{\top}C)$ can be included to control the complexity of the components and avoid rank deficiencies in the eigendecomposition. 

Learning domain-invariant components can be done in a number of ways. The Domain-Invariant Projection approach from \cite{baktashmotlagh2013unsupervised}, later renamed to Distribution-Matching Embedding (DME) \cite{baktashmotlagh2016distribution}, aims to find a projection matrix that minimizes the MMD:
\begin{align}
	\mathrm{D}_{\text{DME}}[M, p_{\mathcal{ S}}, p_{\mathcal{ T}}] = \| \ \mathbb{E}_{\mathcal{ S}} [ \phi(xM)] - \mathbb{E}_{\mathcal{ T}} [ \phi(xM) ] \ \|_{\mathcal{ H}} \, , 
\end{align}
where $M$ is the projection matrix that is being minimized over, with the additional constraint that it remains orthonormal; $M^{\top}M = I$. This constraint is necessary to avoid pathological solutions to the minimization problem. It is possible to add a regularization term that punishes the within-class variance in the domain-invariant space, to encourage class clustering. Alternatively, the same authors have also proposed the same technique, but with the Hellinger distance instead of the MMD \cite{baktashmotlagh2016distribution}. This approach resembles Transfer Component Analysis, but minimizes discrepancy instead of maximizing joint domain variance \cite{pan2011domain}. It has been extended to nonlinear projections as well \cite{baktashmotlagh2016distribution}. 

Alternatively, \cite{muandet2013domain} proposed to learn the values of the MMD kernel itself: instead of weighting or projecting samples and then using a universal kernel to measure their discrepancy, it is also possible to find a basis function for which the two sets of distributions are as similar as possible. The space spanned by this learned kernel then corresponds to the domain-invariant space. Considering that different distributions generate different means in kernel space, it is possible to describe a distribution of kernel means. The variance of this meta-distribution, termed \emph{distributional variance}, should then be minimized to obtain the proposed learned MMD kernel \cite{muandet2013domain}. The functional relationship between the input and the classes can be preserved by incorporating a central subspace in which the input and the classes are conditionally independent \cite{gu2009learning}. Constraining the optimization objective while maintaining this central subspace, ensures that classes remain separable in the new domain-invariant space. This technique is coined Domain-Invariant Component Analysis (DICA) \cite{muandet2013domain}. It has been expanded on for the specific case of spectral kernels by \cite{long2015domain}. Other techniques of this type include information-theoretic learning, where the authors assume that classes across domains are still clustered together in high-dimensional space \cite{shi2012information}.

One could also find the subspace within the data from which the reconstruction to both domains is optimal \cite{si2010bregman,jhuo2012robust}. Transfer Subspace Learning minimizes the Bregman divergence to both domains, with respect to a projection to a subspace \cite{si2010bregman}. In practice, the source data is mapped to a lower-dimensional representation, and then mapped back to the original dimensionality. The reconstruction error then consists of the mismatch between the reconstructed source samples and the target samples, measured through the squared error or the Frobenius norm for instance \cite{shao2014generalized}. 

\subsection{Deep domain adaptation}
Reconstructing the target data from the source data can be done through an autoencoder. Autoencoders are forms of neural networks that reconstruct the input from their hidden layers: $\| \psi(\psi(x M)M^{-1}) - x \|$ where $M$ is a projection matrix and $\psi$ is a nonlinear activation function \cite{hinton1994autoencoders}. There are multiple ways of avoiding trivial solutions to this objective; one formulation will extract meaningful information by pushing the input through a bottleneck (contractive autoencoders) while another adds artificial noise to the input which the network has to remove (denoising autoencoders) \cite{bengio2013RepLearning}. Deep autoencoders can stack multiple layers of nonlinear functions on top of each other to create flexible transformations \cite{vincent2010stacked,glorot2011domain}. Stacking many layers can increase computational cost, but computations for denoising autoencoders can be simplified through noise marginalization \cite{chen2012marginalized}.

Autoencoders are not the only neural networks that have been applied to domain adaptation. One of the most popular adaptive networks is the Domain-Adverserial Neural Network (DANN), which aims to find a representation such that the domains cannot be distinguished from each other while correctly classifying the source samples \cite{ganin2016domain}. It does so by including two loss layers: one loss layer classifies samples based on their labels, while the other loss layer classifies samples based on their domains. During optimization, DANN will minimize label classification loss on while maximizing domain classification loss. Domain-adversarial networks mostly rely on the generalization error bound from Section \ref{sec:geb}: if the domain discrepancy is small, the target error of a source classifier will be small as well. 

Maximizing the classification error of samples from different domains is equivalent to minimizing the proxy ${\cal A}$-distance:
\begin{align}
	\mathrm{D}_{\cal A}[x, z] = 2 \big(1 - 2 \ \hat{e}(x,z) \big) \, ,
\end{align}
where $\hat{e}(x,z)$ is the cross-validation error of a classifier trained to discriminate source samples $x$ from target samples $z$ \cite{kifer2004detecting,ben2010theory}. The proxy ${\cal A}$-distance is derived from the total variation distance \cite{kifer2004detecting}. Other distances have been employed for classifying domains, most notably those based on moment-matching \cite{zellinger2019robust} or using the Wasserstein distance \cite{shen2018wasserstein}. These can be computationally expensive, and there is work on developing less expensive metrics such as the Central Moment Discrepancy or the Paired Hypotheses Discrepancy \cite{zellinger2017central,lee2019domain}.

A limitation of domain-adversarial networks is that matched data distributions do not imply that the class-conditional distributions will be matched as well, as already discussed at the start of this section. Furthermore, the two loss layers produce gradients that are often in different directions. Because of this, DANNs can be harder to train than standard deep neural nets. But recent work has looked at stabilizing the learning process. The DIRT-T approach accelerates gradient descent by incorporating the natural gradient and guiding the network to avoid crossing high-density data regions with its decision boundary \cite{shu2018dirt}. The objective can also be stabilized by replacing the domain-confusion maximizing part of the objective with its dual formulation \cite{usman2017stable}.

The idea of maximizing domain-confusion while minimizing classification error has been explored with numerous network architectures, such as with residual layers \cite{long2016unsupervised}, through generative adversarial networks \cite{bousmalis2017unsupervised}, by tying weights at different levels \cite{tzeng2017adversarial}, embedding kernels in higher layers \cite{long2015learning}, aligning moments \cite{zellinger2019robust} and modeling domain-specific subspaces \cite{bousmalis2016domain}. They have also been applied to a variety of problem settings, such as speech recognition \cite{sun2017unsupervised}, medical image segmentation \cite{kamnitsas2017unsupervised} and cross-language knowledge transfer \cite{huang2013cross}. We refer to \cite{patel2015visual,csurka2017domain,wang2018deep} for extensive reviews.

\subsection{Correspondence learning}
High-dimensional data with correlated features is common in natural language processing as well. In a \emph{bag-of-words} (BoW) encoding, each document is described by a vector consisting of the word occurrence counts. Words that signal each other, tend to co-occur and lead to correlating features. Now, suppose a particular word is a strong indicator of positive or negative sentiment and only occurs in the source domain. One could find a correlating word, referred to as a \emph{pivot} word, that occurs frequently in both domains \cite{blitzer2006domain,louppe2017learning}. Then find the word in the target domain that correlates most with the pivot word. This target domain word is most likely the corresponding word to the original source domain word and will be a good indicator of positive versus negative sentiment as well \cite{blitzer2006domain}. Thus, by augmenting the bag-of-words encoding with pivot words and learning correspondences, it is possible to construct features common to both domains \cite{blitzer2006domain,blitzer2011domain}. Later approaches generalize correspondence learning through kernelization \cite{li2014learning}.

\section{Inference-based approaches} \label{sec:inference}


In this section, we cover methods that incorporate adaptation in the inference procedure. There are many such ways: reformulating the optimization objective, incorporating constraints based on properties of the target domain or incorporating uncertainties through Bayesian inference. As a result, this category is more diverse than the others.

\subsection{Algorithmic robustness} \label{sec:robust}
An algorithm is deemed robust if it can partition a labeled feature space into disjoint sets, or \emph{regions}, such that the variation of the classifier's loss is bounded \cite{xu2012robustness}. This implies that when a training sample is removed from a region, the change in loss is small. Algorithmic robustness naturally extends to the case of data set shift, if one considers the shift to be the addition and removal of samples between training and testing \cite{mansour2014robust}. A classifier that is robust to such changes, can be trained on the source domain and would generalize well to the target domain. 

Changes over posterior probabilities in regions of feature space can be described by $\lambda$-shift. It is said that the target posterior is \emph{$\lambda$-shifted} from the source if the following holds for all $x \in X_r$ and for all $K$ classes: $p_{\cal T}(y = k \mid x) \ \leq \ \ p_{\cal S}(y = k \mid x) + \lambda p_{\cal S}(y \neq k \mid x)$ and $p_{\cal T}(y = k \mid x) \ \geq \  \ p_{\cal S}(y = k \mid x) (1 - \lambda)$
where $X_r$ refers to the $r$-th region. Note that for $\lambda=0$, the posteriors are exactly equal, and for $\lambda = 1$, the posteriors can be arbitrarily different. In essence, this measure is a generalization of the assumption of equal posterior distributions that was used in importance-weighting methods for covariate shift. $\lambda$-shift is a less strict assumption and, hence, algorithms based on $\lambda$-shift are more widely applicable than importance-weighting methods. 

Using the notion of $\lambda$-shift and a robust algorithm, the following holds with probability $1-\delta$, for $\delta > 0$ (Theorem 2, \cite{mansour2014robust}):
\begin{align}
	e_{\cal T}(h) \! - \! \sum_{X_r \in {\cal X}} \hat{p}_{\cal T}(X_r)&\ \ell^{\lambda}(h(x), X_r) \nonumber \\
	& \leq \xi \sqrt{\frac{2 K \log 2 + 2 \log \frac{1}{\delta}}{n}} + \epsilon_{\cal S}  \, . 
\end{align}
where $\xi$ is the upper bound on the loss function, $\epsilon_{\cal S}$ is the bound on the change in loss when a source sample is replaced, $K$ refers to the number of regions and $\hat{p}_{\cal T}(X_r)$ consists of the empirical probability of a target sample falling in region $X_r$. $\ell^{\lambda}(h(x), X_r)$ corresponds to the maximal loss the classifier $h$ incurs with respect to the source samples in the region $X_r$, taking into account the $\lambda$-shift in posterior probability in that region. 


Incorporating $\lambda$-shift into SVM's produces $\lambda$-shift SVM Adaptation ($\lambda$-SVMA) \cite{mansour2014robust}. In its most pessimistic variant, no restrictions are put on the difference between the posterior distributions (i.e. $\lambda$ is set to $1$). Pessimistic $\lambda$-SVMA's dual optimization problem formulation is identical to a standard SVM, except for the addition of an additional constraint: the sum of weights belonging to source samples $x_i$ falling in region $X_r$ has to be less than or equal to the trade-off parameter $\tau$ times the empirical target probability of the $r$-th region; $\sum_{i \in X_r} a_i \leq \tau \hat{p}_{\cal T}(X_r)$. This could be interpreted as that the weights are not allowed to concentrate on regions of feature space where there are few target samples. Robust forms of ridge and lasso regression have been formed \cite{mansour2014robust}, and this approach has been extended to boosting \cite{habrard2016new}. It also resembles another robust SVM approach that aims to find support vectors for the target domain that produce stable labelings \cite{van2017unsupervised}.



\subsection{Minimax estimators} \label{sec:adversarial}
One could view domain adaptation as a classification problem where an adversary changes the properties of the test data set with respect to the training set. Adversarial settings are formalized as minimax optimization problems, where the classifier minimizes risk, with respect to the classifier's parameters, and an adversary maximizes it, with respect to an uncertain quantity \cite{liu2014robust,chen2016robust,kouw2018target}. By working under maximal uncertainty, the classifier adapts in a more conservative manner. 

A straightforward example of such a minimax estimator is the Robust Bias-Aware classifier \cite{liu2014robust}. Its uncertain quantity corresponds to the target domain's posterior distribution. It minimizes risk for one classifier $h$ while the adversary maximizes risk with respect to another classifier $g$. However, given full freedom, the adversary would always produce the opposite classifier and its optimization procedure would not converge. A constraint is imposed, telling the adversary that it needs to pick posteriors that match the moments of the source distribution's feature statistics. The estimator is now written as \cite{liu2014robust}:
\begin{align}
	\hat{h} = \underset{h \in \mathcal{H}}{\arg \min} \ \underset{g \in \mathcal{H} \cap \Xi}{\max} \ \frac{1}{m} \sum_{j=1}^{m} \ell( h(z_j), g(z_j))  \, , 
\end{align}	
where $\Xi$ represents the set of feature statistics that characterize the source domain's distribution and $\mathcal{H} \cap \Xi$ indicates the restriction of the hypothesis class to functions that produce feature statistics equivalent to that of the source domain. This estimator produces a classifier with high-confidence predictions in regions with large probability mass under the source distribution and low-confidence predictions in regions with small source probability mass.

Another minimax estimator, called the Target Contrastive Robust risk estimator \cite{kouw2018target}, focuses on the performance gain that can be achieved with respect to the source classifier. By contrasting the empirical target risk of a new classifier with that of the source classifier, one can effectively exclude classifier parameters that are already known to produce worse risks than that of the source classifier. The estimator is formulated as:
\begin{align}
	\hat{h}_{\cal T} \! = \! \underset{h \in \mathcal{H}}{\arg \min} \ \underset{q}{\max} \ \frac{1}{m} \sum_{j=1}^{m}& \ell \big( h(z_j), q_j \big)  -  \ell \big(\hat{h}_{\cal S}(z_j), q_j \big) , 
\end{align}	
where $\hat{h}_{\cal S}$ is the source classifier and $q$ denotes soft target labels. If no parameters can be found that are guaranteed to perform better than that of the source classifier's, then it will not adapt. It thereby explicitly avoids \emph{negative transfer} \cite{rosenstein2005transfer,wang2019characterizing}. For discriminant analysis models, it can even be shown that the empirical target risk of the TCR estimate is strictly smaller than that of the source classifier, $\hat{R}_{\cal T}(\hat{h}_{\cal T}) < \hat{R}(\hat{h}_{\cal S})$, for the given target samples (Theorem 1,\cite{kouw2018target}). It is hence a transductive classifier. Its weakness is that it can perform poorly if the source classifier is a bad choice for the target domain to begin with.

\begin{figure}[thb]
\centering
\includegraphics[height=110px]{hdis_budapest_age-chol_n256_dboundary.eps}
\includegraphics[height=110px]{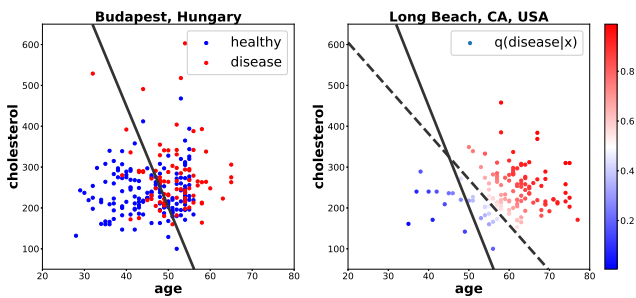}
\caption{Example of a inference-based method. The target samples are soft-labeled through an adversary. The adapted classifier (dashed line) deviates from the unadapted source classifier (solid line) such that the empirical risk on the target samples (right) is lower than that of the source classifier.}
\label{fig:inf-based}
\end{figure}

Figure \ref{fig:inf-based} presents an example of the Target Contrastive Robust risk estimator. By taking into account the uncertainty over the labeling of the target samples, the classifier adapts conservatively (black dashed line). In this case, it is minimizing the empirical risk on the target samples for the worst-case labeling.

An alternative quantity with uncertainty is the set of weights from importance-weighting. Small changes in weights could have large effects on the resulting set of classifier parameters and, by extension, generalization performance. One can be less sensitive to poor weights by minimizing risk with respect to worst-case weights \cite{wen2014robust}:
\begin{align}
	\hat{h} = \underset{h \in \mathcal{H}}{\arg \min} \ \underset{w \in W}{\max} \ \frac{1}{n} \sum_{i=1}^{n} \ell( h(x_i), y_i) w(x_i) \, ,  
\end{align} 
where $W$ is restricted to the set of non-negative weights that average to $1$ (see Section \ref{sec:iw}). 
Minimizing the adversarially weighted loss produces more conservative estimates as the domain dissimilarity increases. Worst-case importance-weights have also been studied from a distributionally robust optimization perspective \cite{hu2018does}. 

Finally, variations on minimax estimators have been proposed in the context of domain-adversarial networks \cite{saito2018maximum}, multiple-kernel support vector machines \cite{sun2013learn} and with using Wasserstein distances \cite{lee2017minimax}.

\subsection{Self-learning} \label{sec:it_self}
A prominent approach in semi-supervised learning is to construct a classifier based on the labeled data and make predictions some of the unlabeled data. These predictions are then treated as labeled data in a following iteration and used for re-training the classifier. This is known as iterative self-labeling or self-learning \cite{mclachlan1975iterative}. A similar procedure can be set up for domain adaptation: train a classifier on the source data, predict the labels of the some of the target samples, and use these labeled target samples to re-train the classifier \cite{perez2007new,bruzzone2010domain}. 

But which target samples to use first? \emph{Co-training} is an approach that uses a logistic regressor to soft-label target samples \cite{chen2011co}. The ones with a posterior probability larger than some threshold, are self-labeled and used in the next iteration of training the classifier. The threshold on the posterior will drop in each iteration, ensuring that the classifier becomes stable.

Another method, called domain adaptation support vector machine (DASVM) \cite{bruzzone2010domain}, will progressively find $2k$ source samples and replace them by $2k$ newly labeled target samples. The $k$ samples closest to the margins spanned by the current support vectors are selected. DASVM has two cost factors that trade-off the source error and the target error. By increasing the cost factor for the target error while decreasing the cost for the source error over time, the classifier focuses increasingly on its performance in the target domain. DASVM has to re-train each iteration, which means it has to solve multiple quadratic optimization problems. That can be computationally expensive. If the data is in the form of strings or trees, these optimization problems can be even more expensive \cite{habrard2013iterative}. The problem in this case is that the underlying kernel has to be positive semi-definite (PSD) and symmetric. By relaxing this constraint and employing similarity functions that need not be PSD or symmetric, a significant speed-up can be achieved \cite{habrard2013iterative}. 

If the target labels are treated as hidden variables, then inference can be done using an Expectation-Maximization procedure \cite{bishop2006pattern,perez2007new}. Adaptation with Randomized Expectation Maximization (Ad-REM) uses the source classification model to compute the expected labeling of the target samples \cite{van2018domain}. Given the expected labels, it maximizes the classification model's likelihood with respect to its parameters. It is hence an iterative self-labeling procedure. Where it differs from previous approaches, such as DASVM, is that it enforces class balance in the target samples: after each expectation step, a roughly equal amount of target samples assigned to each class is taken for the maximization step \cite{van2018domain}. Class balance ensures more stable predictions.

Self-labeling can be combined with a transformation-based technique. Balanced Distribution Adaptation (BDA) uses a source classifier to obtain soft labels for the target samples \cite{wang2017balanced}. These soft labels are then used to estimate the conditional distributions in the target domain, $p_{\cal T}(x \given y)$. In turn, the conditionals are used to match the domains via a linear transformation. After transforming the source domain, the source classifier is re-trained and the target samples are assigned new soft labels. This procedure is re-iterated until convergence.

An alternative procedure is based on PAC-Bayesian weighted majority votes \cite{morvant2015domain,germain2013pac}. First, the notion of \emph{perturbed-variation} (PV) self-labeling is introduced, a measure that returns the proportion of target samples that are not within an $\epsilon$-radius of any source sample \cite{morvant2015domain}. PV is $0$ when all target samples are close to a source sample. The matched target samples, i.e. those within range, are assigned the same label as their closest source sample. This constitutes the self-labeled set, which is later used for estimating the weighted majority vote for an ensemble of classifiers. In total, PV-minCq performs one iteration of self-labeling followed by ensemble training \cite{morvant2015domain}.

Existing domain adaptation approaches can be extended to include self-learning. Joint Distribution Adaptation (JDA) is an extension of Transfer Component Analysis \cite{long2013transfer,pan2011domain}. It pseudo-labels the target samples in order to find transfer components based on both the data marginal distributions and the class-conditional distributions \cite{long2013transfer}. JDA can avoid finding transfer components where the class-conditional distributions differ strongly. Similarly, graph-based matching can be extended to include a pseudo-labeling stage \cite{das2018graph}.

The advantage of a self-labeling approach is that it allows for validation: using the self-labeled target samples as a validation set, the current classifier's hyperparameters can be optimized \cite{bruzzone2010domain}. This is also known as the \emph{reverse validation} strategy \cite{bruzzone2010domain,germain2013pac,morvant2015domain}.

\subsection{Empirical Bayes} \label{sec:Bayesian}
In Bayesian inference, one forms a data likelihood function given a set of parameters and poses a prior distribution over these parameters \cite{gelman2013bayesian}. Once data has been observed, one can derive the posterior distribution of the parameters, and make decisions based on Maximum A Posteriori (MAP) estimation. Crucial to this inference process is the choice of prior distribution. Often, the shape of this prior distribution depends on what the expert believes are suitable values. In the absence of an expert, so-called uninformative priors can be constructed.

In domain adaptation problems, the labeled source domain acts as a form of prior knowledge \cite{raina2006constructing,finkel2009hierarchical,kouw2019cross}. One way to incorporate this knowledge in the inference process, is by fitting the prior distribution to the source data. For example, in NLP classification tasks with bag-of-words feature vectors, we know that features correlate heavily with each other \cite{raina2006constructing}. Suppose a prior distribution is set on parameters of a linear classifier where each feature corresponds to a word. An uninformative prior for the classifier's parameters might look like a diagonal covariance matrix, i.e. no word correlations. But if one has access to a large data set of text documents, e.g. Wikipedia, one could estimate correlations between words. These estimates can replace the covariance matrix of the prior distribution. It is now informative as it contains knowledge gained from the source domain, and can improve performance of the target classifier. Note that this is a form of empirical Bayes as the prior is estimated from data \cite{carlin2010bayes}. The main difference from standard Empirical Bayes is that the data used for fitting the prior originates from a different distribution than the data that will be used for the likelihood function.


In a sense, the source domain informs the model on what kind of feature structure could be expected in the target domain \cite{li2007bayesian}. A limitation of this approach is that, if the domains differ too much, then the informative prior will concentrate on a region of parameter space that is further away from the optimal posterior than the uninformative prior, and produce more uncertain predictions.

Constructing informative priors using the source domain has been done in a number of settings: the recognition of speech utterances of variable length \cite{Shepstone2016TotalVM}, script knowledge induction \cite{Frermann2014AHB}, cross-domain action recognition \cite{bian2012cross}, cross-center brain tissue segmentation \cite{kouw2019cross}, using decision-tree priors \cite{mahmud2009universal,srivastava2013discriminative} and using hierarchical priors \cite{finkel2009hierarchical}.

\subsection{PAC-Bayes} \label{sec:PAC-Bayes}
The PAC-Bayesian framework is a combination of PAC-learning and Bayesian inference \cite{germain2007pac,germain2013pac,morvant2015domain}. One defines a prior distribution $\pi$ over an hypothesis space ${\cal H}$ of classification functions \cite{germain2007pac,germain2013pac}. Through observing data, one aims to obtain a posterior distribution $\rho$ over ${\cal H}$. The expected prediction with respect to hypotheses drawn from $\rho$ is known as the $\rho$-weighted majority vote, or Bayes classifier \cite{germain2007pac,morvant2015domain}. 

PAC-Bayes departs from standard Bayesian inference by replacing the consideration of $\rho$ over the entire hypothesis space by, what is known as, the Gibbs classifier \cite{germain2013pac}. The Gibbs classifier $G_{\rho}$ draws, according to $\rho$, a single hypothesis from ${\cal H}$ to make a prediction.
With it, a domain-adaptive PAC-Bayesian generalization error bound can be derived similar to the one from Section \ref{sec:geb} \cite{germain2013pac}. It contains the source risk of the Gibbs classifier plus a discrepancy term and the error of the ideal-joint-hypothesis. The discrepancy consists of the absolute difference between the disagreements on each domain with respect to $\rho$ \cite{germain2013pac}:
\begin{align}
    \mathrm{D}_{\rho}[p_{\cal T}, p_{\cal S}] = \ \left\vert \ d_{\cal S}(\rho) - d_{\cal T}(\rho) \right\vert \, , 
\end{align}
where $d_{\cal S}(\rho) = \mathbb{E}_{\cal S} \mathbb{E}_{h,h' \sim \rho} \mathds{1}[h(x) \neq h'(x)]$ and likewise for $d_{\cal T}$, are the domain disagreements.


Building on the previous results, a new PAC-Bayesian bound was derived that is not dependent on the error of the ideal joint hypothesis, Instead, it utilizes a term, $\eta_{{\cal T} \setminus {\cal S}}$, describing the risk in the part of the target domain for which the source domain is uninformative
and a factor weighting the source error \cite{germain2016new}:
\begin{align}
    R_{\cal T}(G_{\rho}) -  \beta_{\alpha}(p_{\cal T} \| p_{\cal S}) \big( e_{\cal S}(\rho) \big)^{1- \frac{1}{\alpha}} \! \leq \frac{1}{2} d_{\cal T}(\rho) + \eta_{{\cal T} \setminus {\cal S}} \, . 
\end{align}
This factor is an exponentiated R{\'e}nyi divergence $\log_2 \beta_{\alpha}(p_{\cal T} \| p_{\cal S}) = \frac{\alpha - 1}{\alpha} \mathrm{D}_{\alpha R} [p_{\cal T} \| p_{\cal S} ]$. For $\alpha=2$, it would equal the R{\'e}nyi divergence from the bound for importance weighting (c.f. Equation \ref{sec:geb_iw}) if not for the fact that it is with respect to the joint distribution in each domain instead of the data distributions. This divergence focuses on those parts of feature space where the source domain's support is within the target domain's support. Effectively, the source error has little to no influence in parts of feature space supported by the source domain but not the target domain.

The above bound can be expressed in computable terms by taking a linear Gibbs classifier and assuming a Gaussian distribution over classifier parameters for $\rho$ \cite{germain2016new}. The expected disagreement is bounded by the empirical disagreement and the Kullback-Leibler divergence between the prior $\pi$ and the posterior $\rho$ over the hypothesis space. For spherical Gaussian distributions, the KL-divergence between $\pi$ and $\rho$ constitutes the squared norm of the parameters. Similarly, the expected source error can be upper bounded by the empirical error and, again, the Kullback-Leibler divergence between $\pi$ and $\rho$. The resulting optimization objective, coined Domain Adaptation for Linear Classifiers (DALC), trades off the disagreement of the classifier on the target samples, the classification error on the source samples and the squared norm of the classifier parameters:
\begin{align}
    \hat{\theta} \! = \! \underset{\theta}{\arg \min}\ \tau_1 \sum_{j=1}^{m} \tilde{\Phi} \big(\frac{z_j \theta}{\|z_j\|} \big) \! + \! \tau_2 \sum_{i=1}^{n}\Phi^{2} \big(\frac{y_i x_i \theta}{\|x_i\|} \big) \! + \! \| \theta \|^2 ,
\end{align}
where $\Phi$ is a probit loss function and $\tilde{\Phi}(x) = 2\Phi(-x)\Phi(x)$. $\tau_1$ and $\tau_2$ are trade-off parameters based on normalization factors. DALC resembles a standard linear classifier in that it minimizes an average loss. However, it includes an additional term that penalizes predictions on the target samples that are far removed from zero.

\section{Discussion} \label{sec:discussion}
We discuss insights, limitations and points of attention.

\subsection{Assumptions, tests \& no-free-lunch}
The problem with assumptions on the relationship between domains, such as covariate or prior shift, is that they cannot be verified without target labels. In order to check that the posteriors in each domain are actually equal, one would need labeled data from each domain. 
The inability to check for the validity of assumptions means that it is impossible to predict how well a method will perform on a given data set. For any adaptive classifier, it is possible to find a case where learning fails dramatically. Clearly, in domain adaptation there is no free lunch either \cite{wolpert1996lack}. 

Advising practitioners on classifier selection is therefore difficult. Ideally, there would be some hypothesis test that gives an impression of how likely an assumption is to be valid for a given problem setting. A number of such tests have been developed for domain adaptation procedures. Firstly, a test exists for regression under sample selection bias \cite{melino1982testing}. Secondly, the necessity of data importance-weighting can be tested by comparing the adversarial loss of the classifier parameters obtained by training \emph{with} versus \emph{without} data importance-weights, where the adversarial loss consists of maximizing loss given fixed weights \cite{white1981consequences,wen2014robust}. The larger the difference, the more the model is mis-specified and the larger the necessity for importance-weighting. Thirdly, testing for label shift can be done through a two-sample test comparing the empirical source prior and the empirical predicted target prior, as it can be shown that, under weak assumptions, $\hat{p}_{\cal T}(y) = \hat{p}_{\cal S}(y)$ if and only if $p_{\cal T}(y) = p_{\cal S}(y)$ \cite{lipton2018detecting}. 
Lastly, in some cases, a large empirical discrepancy can indicate that no method has the capacity to generalize well to the target domain \cite{cortes2014domain}. 

Given the difficulty of domain adaptation \cite{ben2012hardness}, these tests are crucial to developing practical solutions. Without some indication that your proposed method relies on a valid assumption, the risk of negative transfer is very real \cite{wang2019characterizing}.

\subsection{Insights \& interpretability}
We argue that interpretability, in the sense that one can easily inspect the inner mechanics of a method and gain some intuition on why it is likely to succeed or fail, is an important property. Interpretations lead to novel insights, which deepen our understanding of domain adaptation.

In models with explicit descriptions of transfer, typical success and failure cases can be compared and interpreted. For example, in sample-based methods, the importance weights describe transfer because they indicate how much correction is necessary. In successful adaptation, the weights tend to vary smoothly around the value one. In one set of failure cases, we observe a bimodal distribution with a small set of weights having large values and the remainder being close to zero. Upon reflection, we realize that the adaptive classifier under-performs because the effective sample size drops for such a bi-modal weight distribution.
Similarly, in transformation-based methods, the transformation function itself describes transfer. In success cases, there are few transformations that fit well. 
If many transformations fit, such as when the class-conditional distributions overlap substantially in each domain, then chances are low that the adaptive classifier recovers the correct one. Lastly, some of the inference-based methods fail when facing problems with substantial target probability mass outside the support of the source data distribution. This can be observed through the adversarial loss becoming large or through the disagreement of the learned prior with the posterior relative to an uninformative prior.

\subsection{Shrinking search space}
In a number of methods the source domain is considered a means of narrowing the search space. For example, in empirical Bayes approaches, the informed prior will assign low probabilities to parameters that would not be suitable for the given task. Hence, the search space is effectively smaller than for the uninformative prior. Another example is \emph{fine-tuning}, where a neural network is first trained on a large generic data set and then trained on the smaller data set of the problem of interest. Starting with more plausible parameters than random ones means that it will avoid many implausible parameters \cite{mcnamara2017risk}. 

Ignoring parts of parameter space is an intuitive notion of how the source domain assists the learning process. This interpretation shows an interesting link to natural intelligence, where agents use knowledge from previous tasks to complete new tasks \cite{kanazawa2004general}. But ignoring parameters can be dangerous as well. If parameters are ignored that \emph{are} actually useful for the target domain, then the source domain effectively interferes with learning. 


%

\subsection{Multi-site studies}
It is notoriously difficult to integrate data from research groups working in the same field. The argument is that another group uses different experimental protocols or measuring devices, or is located in a different environment, and that their data is therefore not compatible \cite{leek2010tackling}. For example, in biostatistics, gene expression micro-array data can exhibit \emph{batch effects} \cite{johnson2007adjusting}. These can be caused by the amplification reagent, time of day, or even atmospheric ozone level \cite{fare2003effects}. In some data sets, batch effects are actually the most dominant source of variation and are easily identified by clustering algorithms \cite{johnson2007adjusting}. 

Domain adaptation methods are useful tools for this type of problem. Additional information such as local weather, laboratory conditions, or experimental protocol, can be exploited to correct for the data shift. 
Considering the financial costs of some types of experiments, the ability to remove batch effects and integrate data sets from multiple research centers would be valuable. 
Proper cross-domain generalization techniques could indirectly increase the incentive for making data sets publicly available.

\subsection{Causality}
Domain shifts can be viewed as the effect of one or more variables on the data-generating process \cite{scholkopf2012causal,zhang2013domain}. Knowledge of how those variables affect features can clarify whether assumptions, such as covariate shift or prior shift, are reasonable to make \cite{gong2016domain,rojas2018invariant}. Given the causal structure, one could even relax the covariate shift assumption to hold only for a subset of the features \cite{rojas2018invariant}. Furthermore, taking the causal relationships between variables into account can lead to better predictions of invariant conditional distributions\cite{zhang2013domain,magliacane2018domain}. For example, recent work considers the domains to be context variables outside of the system of interest \cite{magliacane2018domain}. By making certain assumptions on the confounding nature of these context variables, one could identify a set of features that is invariant under transfer from the source to target domain. These assumptions on context variables can often be linked to real-world environmental variables, such as the organism in a genetics study or the environment in a data-collecting research institute \cite{magliacane2018domain,zhang2015multi}. The main limitation of causal inference procedures is that they, currently, do not scale beyond dozens of variables. As the causal structure resolves ambiguities on the type of domain shift occurring in any given problem, advances in causal discovery could lead to automatic adaptation strategy selection and eventually, to much more practical domain-adaptive systems.


\subsection{Limitations}
Methods have general limitations. Firstly, importance-weighting suffers from the curse of dimensionality. The weights are supposed to align distributions, but measuring alignment in high-dimensional settings is tricky. Secondly, a variety of methods require that the support of the target distribution is contained within the support of the source distribution, where ''support'' should be interpreted as the collection of areas of high probability mass. If this is not the case, then pathological solutions such as high weight variance or worse-than-uninformative priors can occur. Thirdly, unfortunately, combining transformations that bring the supports of the distributions closer together with methods that involve support requirements, is not straightforward: subjecting the source and target domain to different transformations can break the equivalence of posterior or conditional distributions. Fourthly, in many practical high-dimensional settings, there are a multitude of transformations that would align the data distributions but not the posterior or class-conditional distributions. Heuristics and knowledge of the particular application of interest is required to pick an appropriate transformation.

Our survey is limited in that it is not exhaustive. This is due to two reasons: firstly, we selected papers with the goal of presenting what we believe are original ideas, thereby disregarding special cases and particular applications. Secondly, inconsistent terminology in the literature makes it hard to find papers that adhere to our problem definition but follow alternative naming conventions. Some such articles have been found by chance and are included here, such as \cite{si2010bregman}, but we cannot exclude the possibility that some have escaped our search criteria. Nonetheless, we do believe to have presented the reader a thorough summary.

\subsection{Terminology}
The literature contains a wide variety of terminology describing the current problem setting. Our setting "domain adaptation without target labels" has been referred to as "unsupervised domain adaptation", "transductive transfer learning", and terms involving sampling bias, correcting distributional shifts, robustness to data shifts or out-of-sample / out-of-distribution generalization. 
We tried to be consistent in our terminology, but the reader should be aware that, in general, the literature is not.

\section{Conclusion} \label{sec:conclusion}
We reviewed work in domain adaptation to answer the question: \emph{how} can a classifier learn from a source and generalize to a target domain? 
We have made a coarse-level categorization into three categories: those that operate on individual observations (sample-based), those that operate on the representation of sets of observations (feature-based) and those that operate on the parameter estimator (inference-based). On a finer level, we can split sample-based methods into data importance-weighting, based on assuming covariate shift, and class importance-weighting, based on assuming prior shift. A variety of weight estimators has been proposed, suitable to diverse problem settings and data types. Similarly, we can split feature-based methods into subspace mappings, optimal transport, domain-invariant spaces, deep domain adaptation and correspondence learning. Subspace mappings focus on transformations from source to target based on subsets of feature space. Optimal transport also focuses on transformations from source to target, but does so on the level of probability distributions. Domain-invariant representations, where both source and target data are mapped to a new space, can be learned. Deep domain adaptation is the neural network equivalent of that. Correspondence learning constructs common features through feature inter-dependencies in each domain. As can be imagined, these methods require high-dimensional data, such as images or text. Subspace projections are more common to computer vision while learning domain-invariant representations is popular in both the image and the natural language processing community. Inference-based methods is a diverse category consisting of algorithmic robustness, minimax estimators, self-learning, empirical Bayes and PAC-Bayes. Robust algorithms aim to partition feature space such that the loss is small when removing a training sample and minimax estimators optimize for worst-case shifts under constraints imposed on the adversary. Self-learning procedures provisionally label target samples and include the pseudo-labeled samples in training the target classifier. In empirical Bayes, the prior distribution is fit to the source data and in PAC-Bayes, one considers disagreement between hypotheses drawn from the posterior over the hypothesis space while ignoring source samples lying outside the support of the target data distribution.

Furthermore, we find that assumptions are a necessary component to domain adaptation without target labels and that these strongly influence when a particular method will succeed or fail. It is crucial to develop hypothesis tests for the validity of such assumptions. Moreover, to develop domain adaptation methods, it is important to study interpretable procedures. Comparing explicit descriptions of transfer between success and failure cases can produce novel insights. Additionally, domain adaptation is not only relevant to many scientific and engineering disciplines, it is of value to integrating multi-site data sets and to the computational expense of existing algorithms. Automatically discovering causal structure in data is an exciting possibility that could resolve many ambiguities on adaptive classifier selection. In general, domain adaptation is an important problem, as it explores outside the standard assumptions of independently and identically distributed data. It should be studied in further detail.

\ifCLASSOPTIONcaptionsoff
  \newpage
\fi



%
\bibliographystyle{IEEEtran}
\bibliography{kouw_pami18a}

\vfill


\end{document}